\documentclass[sigconf]{acmart}
\AtBeginDocument{%
	}

\author{Zeyu Zhang$^1$, Yang Zhang$^2$, Haoran Tan$^1$, Rui Li$^1$, Xu Chen$^1$}
\affiliation{%
	\institution{$^1$Gaoling School of Artificial Intelligence, Renmin University of China, China}
	\country{}
}
\affiliation{%
	\institution{$^2$National University of Singapore, Singapore}
	\country{}
}
\email{{zeyuzhang, xu.chen}@ruc.edu.cn, zyang1580@gmail.com}

\usepackage{bigstrut}
\usepackage{xcolor}
\usepackage{pifont}
\usepackage{graphicx}
\usepackage{subcaption}
\usepackage{tcolorbox}
\usepackage{titlesec}
\usepackage{colortbl}
\usepackage{multirow}

\theoremstyle{definition}
\newtheorem{definition}{Definition}

\definecolor{darkgreen}{RGB}{0,100,0}
\definecolor{nr-blue}{HTML}{006699}
\definecolor{sr-red}{HTML}{CC0033}
\definecolor{mr-green}{HTML}{669933}
\definecolor{dr-purple}{HTML}{996699}

\titlespacing*{\section}{0pt}{3pt}{2.5pt}

\titlespacing*{\subsection}{0pt}{3pt}{2.5pt}

\titlespacing*{\subsubsection}{0pt}{2.5pt}{2.5pt}

\begin{document}
	
\title{Explicit v.s. Implicit Memory: Exploring Multi-hop Complex Reasoning Over Personalized Information}

\begin{abstract}
In large language model-based agents, memory serves as a critical capability for achieving personalization by storing and utilizing users' information.
Although some previous studies have adopted memory to implement user personalization, they typically focus on preference alignment and simple question-answering.
However, in the real world, complex tasks often require multi-hop reasoning on a large amount of user information, which poses significant challenges for current memory approaches.
To address this limitation, we propose the multi-hop personalized reasoning task to explore how different memory mechanisms perform in multi-hop reasoning over personalized information.
We explicitly define this task and construct a dataset along with a unified evaluation framework.
Then, we implement various explicit and implicit memory methods and conduct comprehensive experiments. We evaluate their performance on this task from multiple perspectives and analyze their strengths and weaknesses.
Besides, we explore hybrid approaches that combine both paradigms and propose the HybridMem method to address their limitations.
We demonstrate the effectiveness of our proposed model through extensive experiments.
To benefit the research community, we release this project at \url{https://github.com/nuster1128/MPR}.
\end{abstract}


\maketitle

\section{Introduction}

With the rapid advancement of large language models (LLMs)~\cite{zhao2023survey,chang2024survey,naveed2023comprehensive}, LLM-based agents have been widely applied for providing personalized services~\cite{wang2024survey,xi2025rise,guo2024large}.
Specifically, memory serves as a key component for achieving personalized agents, responsible for storing and utilizing personalized information to meet users' requirements~\cite{zhang2024survey}.
Although previous works have proposed various memory approaches to improve personalization, most of them focus on user preference alignment tasks~\cite{salemi2023lamp,tan2024democratizing} and simple question-answering~\cite{du2024perltqa}.
These tasks do not demand explicit reasoning processes by aggregating users' factual information.
However, in real-world personalized applications, complex tasks that require multi-hop reasoning over a large amount of personalized information are more practical and challenging for the memory of agents, which still remains unexplored.
To fill this gap, we focus on multi-hop personalized reasoning (MPR) tasks in this study.

\begin{figure*}[t]
	\centering
	\setlength{\fboxrule}{0.pt}
	\setlength{\fboxsep}{0.pt}
	\vspace{-0.1cm}
	\fbox{
		\includegraphics[width=1.0\linewidth]{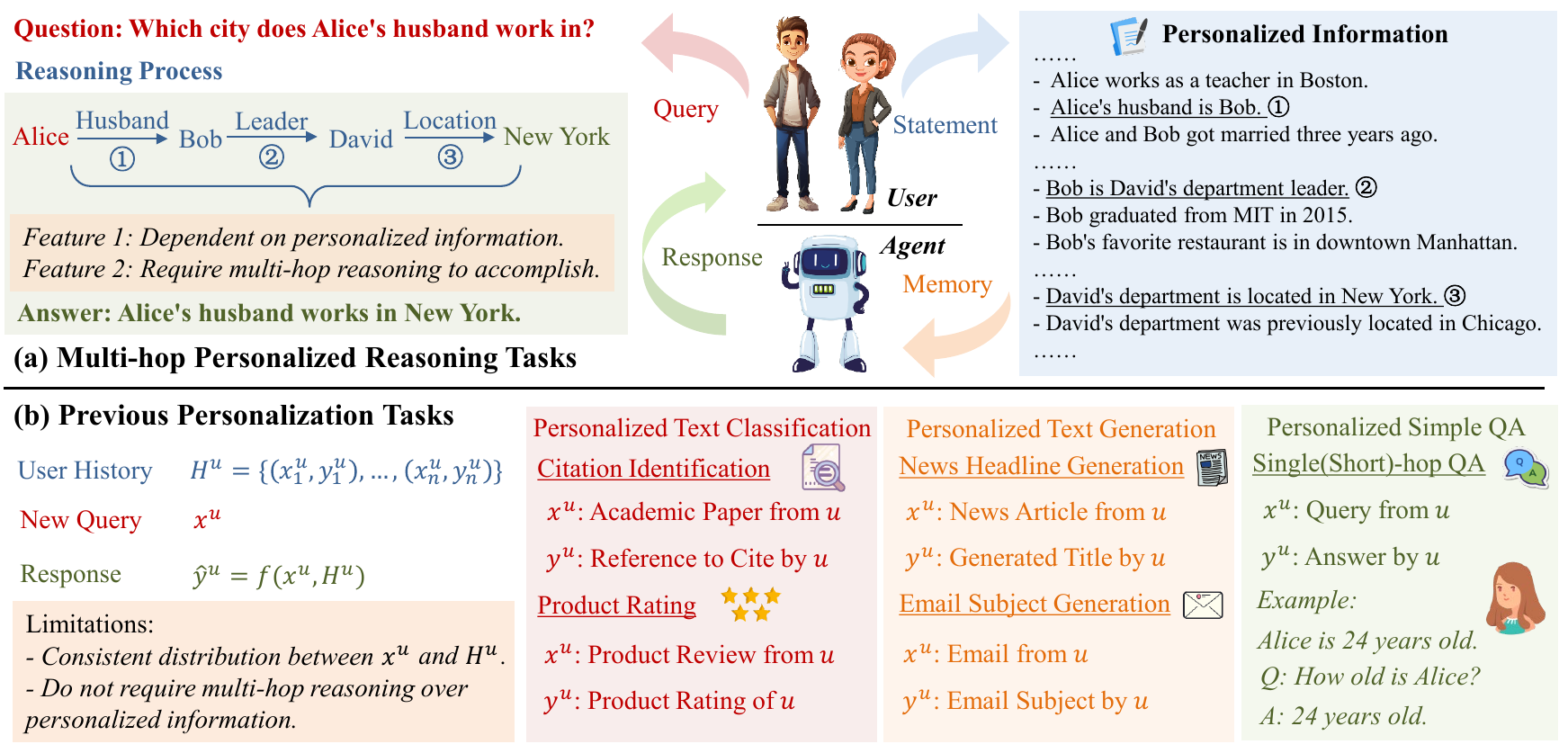}
	}
	\vspace{-0.6cm}
	\caption{Demonstration of multi-hop personalized reasoning tasks and previous personalization tasks.}
	\label{fig:introduction}
	\vspace{-0.4cm}
\end{figure*}

MPR tasks are characterized by two key features: (1) tasks must be completed based on given personalized information and cannot be accomplished solely through general knowledge; (2) tasks can not be directly accomplished with a single piece of personalized information, but require multi-hop reasoning on several pieces.
We demonstrate an example of MPR tasks in \textbf{Figure~\ref{fig:introduction}(a)}, and a formal definition will be provided in later sections.

We emphasize that MPR tasks are significantly different from previous personalization tasks, as illustrated in \textbf{Figure~\ref{fig:introduction}(b)}.
First, most previous tasks keep consistent distributions between training data (\textit{i.e.,} user histories) and testing data (\textit{i.e.,} predictions) that are drawn from the same distribution of user utterances~\cite{salemi2023lamp}.
However, each MPR task requires multiple pieces of user information, leading to composition gaps between training and testing.
Besides, previous tasks do not emphasize reasoning processes, whereas MPR tasks necessitate multi-hop reasoning over massive personalized information to accomplish.
Finally, user histories in MPR tasks consist of factual information for complex reasoning, rather than preference or stylistic information~\cite{salemi2023lamp, kumar2024longlamp}.

Like other personalization tasks, users' personalized information should be memorized by LLM-based agents to accomplish MPR tasks.
According to previous works, agent memory can be categorized into explicit and implicit forms based on their representation~\cite{zhang2024survey}.
Explicit memory refers to storing users' personalized information in textual form and utilizing it through in-context learning (ICL)~\cite{dong2022survey}, typically implemented through Retrieval-Augmented Generation (RAG) methods~\cite{fan2024survey}.
Implicit memory refers to storing users' personalized information within model parameters, typically implemented through supervised fine-tuning (SFT) methods~\cite{zhang2024instruction}.
However, considering the multi-hop reasoning feature of MPR tasks, explicit memory may suffer from retrieval mismatch problems between reasoning steps, while implicit memory may face difficulties in accurately storing large amounts of factual personalized information. Therefore, how explicit and implicit memory perform on MPR tasks represents a valuable but unexplored question.

In this paper, we focus on exploring explicit and implicit memory on MPR tasks.
We formally define MPR tasks and construct datasets along with an evaluation framework for memory mechanisms in multi-hop reasoning.
We comprehensively study explicit memory, implicit memory, and hybrid memory on these tasks from multiple perspectives, drawing several key findings.
We find that explicit memory demonstrates clear advantages across various reasoning structures, and incorporating implicit memory can enhance explicit memory.
Additionally, we discover that reasoning structure significantly affects performance across all memory types.
Finally, we propose a simple yet effective hybrid memory approach to improve the performance of agent reasoning on long-hop tasks.
To benefit the research community, we release our code at the GitHub repository \url{https://github.com/nuster1128/MPR}.

Our major contributions are summarized as follows: \\
$\bullet$ We identify the MPR tasks, highlighting their unique challenges for agent memory compared to previous works.\\
$\bullet$ We formally define MPR tasks and construct a new dataset with a unified evaluation framework for systematically exploring different memory methods on MPR tasks.\\
$\bullet$ We conduct comprehensive experiments on explicit, implicit, and hybrid memory approaches, presenting key findings and proposing a new hybrid memory method for long-hop reasoning tasks.

\section{Related Work}

\subsection{Personalized LLM-based Agents}
Recently, LLM-based agents have been extensively applied to user personalization tasks for enhancing user experiences. Many previous works focus on aligning LLMs with user preferences to achieve user personalization~\cite{li2024personal}.
For instance, LaMP~\cite{salemi2023lamp} constructs a personalized benchmark to evaluate models' ability to infer subsequent user responses given user historical utterances.
CFRAG~\cite{shi2025retrieval} incorporates users' previous documents in textual form through RAG along with collaborative information to improve personalization.
OPPU~\cite{tan2024democratizing} fine-tunes a LoRA adapter for each user based on previous utterances to empower personalization.
Besides, some studies achieve agent personalization by delivering personalized services.
For example, PerLTQA~\cite{du2024perltqa} constructs personalized questions to evaluate agents' long-term memory and proposes a retrieval-based memory framework.
However, these tasks do not consider the multi-hop reasoning complexity in real-world tasks, and pay less attention to the distributional mismatch between personalized information and tasks.
Therefore, in this paper, we intend to highlight the importance of exploring MPR tasks.


\subsection{Memory of LLM-based Agents}
Memory is a crucial capability of LLM-based agents, responsible for storing historical information and utilizing relevant content to support decision-making~\cite{zhang2024survey}.
It is typically categorized into explicit and implicit forms.
Explicit memory stores and utilizes information in textual form. For example, MemoryBank~\cite{zhong2024memorybank} stores past conversations as text and retrieves semantically relevant information using retrieval models.
In contrast, implicit memory stores and utilizes information within model parameters. For instance, MEND~\cite{mitchell2021fast} leverages meta-learning to train a model that transforms knowledge into parameter adjustments.
In addition, some previous works also focus on agent memory evaluation~\cite{zhang2024memsim,tan2025membench,wu2024longmemeval} and development~\cite{zhang2025memengine, chhikara2025mem0}, particularly in long-term scenarios.
Agents can accomplish user modeling by designing memory mechanisms to capture users' preferences~\cite{zhang2024survey}. This enables the personalized agents to avoid the "one-size-fits-all" phenomenon~\cite{chen2024large}.


\subsection{Multi-hop Reasoning in Agents}
Reasoning is important for LLM-based agents to accomplish complex tasks, enabling them to decompose one task into multiple steps, thereby reducing the difficulty of inference~\cite{huang2024understanding}.
Chain-of-thought (CoT)~\cite{wei2022chain} is a representative work that enhances the reasoning capabilities of LLMs by adding instructions to make the model think step-by-step.
After that, many different structures have been proposed to implement reasoning processes, such as Tree-of-Thought (ToT)~\cite{yao2023tree} and divide-and-conquer approaches~\cite{wang2023plan}.
Some previous works have also constructed multi-hop datasets to evaluate agents' reasoning on general tasks~\cite{schnitzler2024morehopqa,trivedi2022musique,zhu2024fanoutqa}.
We highlight that MPR tasks are different from previous open-ended reasoning tasks.
First, the information in MPR tasks is user-specific and cannot be obtained in advance (\textit{e.g.,} from Wikipedia), avoiding reasoning shortcuts.
Moreover, their difficulties depend not only on the problems themselves, but also on the available reasoning evidence, which is controllable.


\section{Preliminaries}

\subsection{Problem Definition}
First of all, we deliver a formal definition of MPR tasks as follows:

\begin{definition}[MPR Task]
	\vspace{-0.1cm}
	Given a collection of personalized statements $\mathcal{S}=\{s_1, s_2, ..., s_n\}$, where $s_i$ ($1 \leq i \leq n$) describes a single-hop factual information from the user, the model $f$ is required to predict an answer $\hat{a}=f(q;\mathcal{S})$ to a question $q$ based on $\mathcal{S}$, in order to match the correct answer $a$. Meanwhile, the tasks should satisfy:\\
	(1) $\exists S \subseteq \mathcal{S} (S \neq \emptyset)$ such that $S$ is necessary and sufficient to infer the correct answer $a$ to question $q$.\\
	(2) $\nexists s_i \in \mathcal{S}$ such that the correct answer $a$ to question $q$ can be obtained solely from $s_i$ without other statements.
	\vspace{-0.1cm}
\end{definition}
These two conditions correspond to the features of MPR tasks and serve as important criteria that distinguish them from other tasks.
For instance, for the question "\textit{Which city does Alice's husband work in?}" in \textbf{Figure~\ref{fig:introduction}(a)}, we can only obtain the answer from the user's statement, rather than relying on general knowledge.
Furthermore, the user's statement does not contain a direct answer such as "\textit{Alice's husband works in New York.}"
Instead, the answer must be derived through reasoning across multiple statements.

\subsection{Solution Paradigm}
\label{sec:solution_paradigm}
The most naive approach for MPR tasks is to integrate the entire statement collection $\mathcal{S}$ and question $q$ into a prompt, and let the LLM directly infer the answer, that is,
$$
\setlength\abovedisplayskip{2pt}
\setlength\belowdisplayskip{2pt}
\hat{a} = f(q;\mathcal{S}) = g(s_1 \| s_2 \| ... \| s_n \| q; \theta),
$$
where the function $g(x;\theta)$ represents an LLM inference for prompt $x$ with parameter $\theta$, and $\|$ denotes string concatenation.
However, this naive approach has two obvious limitations. First, due to the large scale of users' personalized information, it is almost impossible to integrate all these statements in the prompt.
Therefore, one solution is to rely on RAG to select a subset $\hat{S} \subseteq \mathcal{S}$ as the explicit memory, and we have
\setlength\abovedisplayskip{1pt}
\setlength\belowdisplayskip{1pt}
\begin{gather*}
	\hat{a} = f(q;\mathcal{S}) = g(\hat{s}_1 \| \hat{s}_2 \| ... \| \hat{s}_k \| q; \theta),\\
	\hat{S} = \{\hat{s}_1, \hat{s}_2, ..., \hat{s}_k\} = \text{RAG}(q, \mathcal{S}),
\end{gather*}
where $k$ is the number of retrieved statements.
Another approach is to convert the information into parameter modifications as implicit memory through SFT, and we have
\setlength\abovedisplayskip{1pt}
\setlength\belowdisplayskip{1pt}
\begin{gather*}
	\hat{a} = f(q;\mathcal{S}) = g(q; \hat{\theta}), \\
	\hat{\theta} = \text{SFT}(\theta; \mathcal{S}),
\end{gather*}
where $\hat{\theta}$ is a personalized parameter under $\mathcal{S}$.

Another problem is that question $q$ requires multi-hop reasoning to solve, so single-step inference $g(x;\theta)$ of LLMs may not be able to predict the answer effectively.
Therefore, it is necessary to implement question processing and answering with multi-hop reasoning structures. 
CoT is a commonly used approach for multi-hop reasoning. Formally, we have the following iterative processes:
\setlength\abovedisplayskip{1pt}
\setlength\belowdisplayskip{1pt}
\begin{gather*}
	o_1 = g_1(s_1 \| s_2 \| ... \| s_n \| q;\theta),\\
	o_i = g_i(s_1 \| s_2 \| ... \| s_n \| q \| o_1 \| o_2 \| ... \| o_{i-1} ;\theta), 2 \le i \le l-1,\\
	\hat{a} = f(q;\mathcal{S}) = g(s_1 \| s_2 \| ... \| s_n \| q \| o_1 \| o_2 \| ... \| o_{l-1} ;\theta),
\end{gather*}
and can be replaced by explicit memory and implicit memory:
\setlength\abovedisplayskip{2pt}
\setlength\belowdisplayskip{2pt}
\begin{gather*}
	\hat{S}_i = \{\hat{s}_1^i, \hat{s}_2^i, ..., \hat{s}_k^i\} = \text{RAG}(o_1 \| o_2 \| ... \| o_{i-1} \| q, \mathcal{S}),\\
	\hat{\theta} = \text{SFT}(\theta; \mathcal{S}).
\end{gather*}
Therefore, the memory problem and reasoning problem correspond to the two major challenges of MPR tasks, that is, how to store and retrieve knowledge, and how to use knowledge for inference.
They are also the core focus of our subsequent experimental exploration.


\begin{table}[t]
	\centering
	\caption{The comparison of related datasets. KR: knowledge-intensive reasoning. PK: private knowledge. PE: personalized evidence of factual information. SR: supporting reference. EC: explicit chain of reasoning.}
	\vspace{-0.4cm}
	\resizebox{\linewidth}{!}{
		\begin{tabular}{cccccccc}
			\hline
			\hline
			\textbf{Dataset} & \textbf{\# Hop} & \textbf{KR} & \textbf{PK} & \textbf{PE} & \textbf{SR} & \textbf{EC} & \textbf{\# QA} \bigstrut\\
			\hline
			MoreHopQA~\cite{schnitzler2024morehopqa} & $\le$ 5 & \ding{51} & \ding{55} & \ding{55} & \ding{51} & \ding{51} & 3,620 \bigstrut[t]\\
			HybridQA~\cite{chen2020hybridqa} & $\le$ 5 & \ding{51} & \ding{55} & \ding{55} & \ding{51} & \ding{55} & 69,611 \\
			2WikiMultiHopQA~\cite{ho2020constructing} & $\le$ 5 & \ding{51} & \ding{55} & \ding{55} & \ding{51} & \ding{51} & 192,606 \\
			MuSiQue~\cite{trivedi2022musique} & $\le$ 5 & \ding{51} & \ding{55} & \ding{55} & \ding{51} & \ding{51} & 24,814 \\
			FanOutQA~\cite{zhu2024fanoutqa} & $\sim$ 10 & \ding{51} & \ding{55} & \ding{55} & \ding{51} & \ding{51} & 1,035 \\
			HotpotQA~\cite{yang2018hotpotqa} & $\le$ 5 & \ding{51} & \ding{55} & \ding{55} & \ding{51} & \ding{55} & 112,779 \\
			MultiHop-RAG~\cite{tang2024multihop} & $\le$ 5 & \ding{51} & \ding{55} & \ding{55} & \ding{51} & \ding{55} & 2,556 \\
			PerLTQA~\cite{du2024perltqa} & $\le$ 5 & \ding{51} & \ding{51} & \ding{55} & \ding{51} & \ding{55} & 8,593 \\
			LongMemEval~\cite{wu2024longmemeval} & $\le$ 5 & \ding{51} & \ding{51} & \ding{55} & \ding{51} & \ding{55} & 500 \\
			CoFCA~\cite{wu2024cofca} & $\le$ 5 & \ding{51} & \ding{51} & \ding{55} & \ding{51} & \ding{51} & 4,500 \\
			MQuAKe~\cite{zhong2023mquake} & $\le$ 5 & \ding{51} & \ding{51} & \ding{55} & \ding{51} & \ding{51} & 9,218 \\
			MPR (Ours) & $\sim$ 10 & \ding{51} & \ding{51} & \ding{51} & \ding{51} & \ding{51} & 108,000 \bigstrut[b]\\
			\hline
			\hline
	\end{tabular}}
	\vspace{-0.6cm}
	\label{tab:reletive_datasets}
\end{table}%

\subsection{Dataset Construction}
\subsubsection{Previous Datasets and Benchmarks.}
There is still a lack of comprehensive datasets to evaluate MPR tasks.
Some multi-hop question-answering (QA) datasets provide reasoning problems, but most of them rely on public knowledge such as Wikipedia, rather than personalized information~\cite{yang2018hotpotqa}. This poses risks of reasoning shortcuts and information leakage in pre-training corpora.
Although some datasets construct personalized information for QAs, their questions typically contain 1 or 2 hops~\cite{du2024perltqa}.
They do not provide explicit reasoning chains, and the number of QAs is limited.
Therefore, we propose to construct a new MPR dataset to better evaluate MPR tasks, which contains 108,000 personalized reasoning tasks ranging from 2 to 10 hops, and provides explicit reasoning paths with reference evidence.
We compare our MPR Dataset with previous datasets from multiple perspectives in \textbf{Table~\ref{tab:reletive_datasets}}.


\subsubsection{MPR Dataset Generation.}
\begin{figure*}[t]
	\centering
	\setlength{\fboxrule}{0.pt}
	\setlength{\fboxsep}{0.pt}
	\fbox{
		\includegraphics[width=1.0\linewidth]{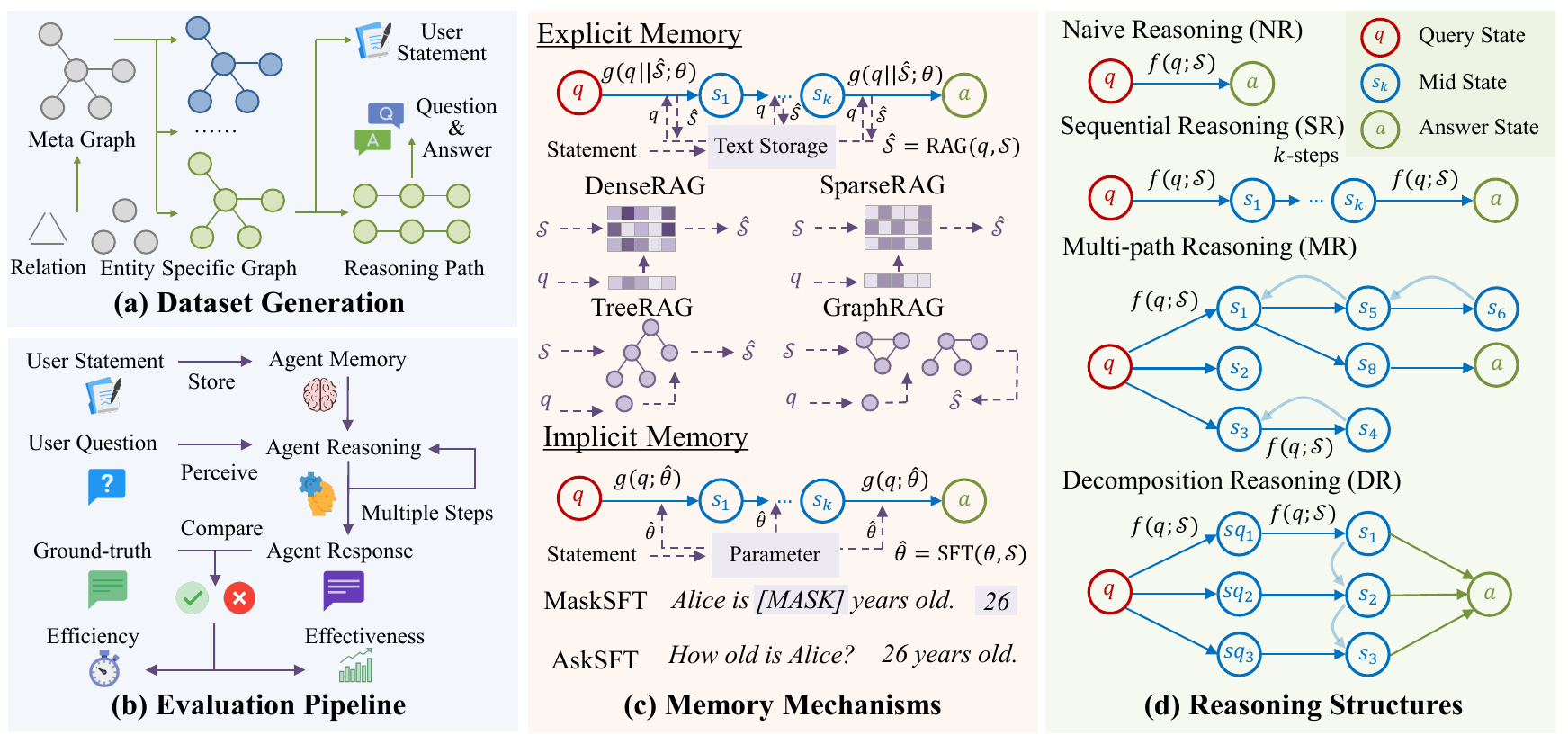}
	}
	\vspace{-0.7cm}
	\caption{Overview of exploration on multi-hop personalized reasoning tasks.}
	\label{fig:pipeline}
	\vspace{-0.4cm}
\end{figure*}

First, we construct a meta graph to represent the graph space before instantiating specific graphs.
Then, we generate user statements from edges, and construct multi-hop questions through path sampling on the graph, with answers assigned by path endpoints.
The process is presented in \textbf{Figure~\ref{fig:pipeline}(a)}.

\textbf{Step 1: Generate Meta Graph.}
In order to improve user diversity, we define a meta graph as the sampling space for users.
Formally, we denote the meta graph as $\mathcal{G} = (\mathcal{V}, \mathcal{E})$, where $\mathcal{V}$ and $\mathcal{E}$ represent the node set and edge set respectively. 
In meta graph $\mathcal{G}$, each node $V_i \in \mathcal{V} (1 \le i \le |\mathcal{V}|)$ represents an conceptual space for a type of entity or attribute (\textit{e.g.,} persons), containing specific entities or attribute values (\textit{e.g.,} Alice) that can be sampled, that is, $V_i = \{v_1^i, v_2^i, ..., v_{|V_i|}^i\}$.
Each edge $E_i \in \mathcal{E} (1 \le i \le |\mathcal{E}|)$ represents a relation space (\textit{e.g.,} social relations) between two nodes, including specific relationships (\textit{e.g.,} be supervised by) that can be instantiated, that is, $E_i = \{e_1^i, e_2^i, ..., e_{|E_i|}^i\}$.
According to the type of two connected nodes (entity or attribute), we categorize all these edges into three categories. 
Entity-oriented edges connect two entities, representing the relationship between them.
Attribute-oriented edges connect an entity and an attribute, indicating that the entity possesses a value of this attribute.
Value-oriented edges connect two attributes, representing the comparative relationship between them.
The first two types of edges involve reasoning with personalized information, while the last type relies on general knowledge, which is designed to enhance the diversity of reasoning.

\textbf{Step 2: Instantiate Specific Graphs.}
Then, we sample attribute values of nodes using LLMs, and simultaneously instantiate edges. 
We organize all the sampled nodes $\tilde{V} = \{ \tilde{v}_i | \tilde{v}_i \in V_i, 1 \le i \le |\mathcal{V}| \}$ and edges $\tilde{E} = \{\tilde{e}_i | \tilde{e}_i \in E_i, 1 \le i \le |\mathcal{E}| \}$ as a specific graph $\tilde{G} = (\tilde{V}, \tilde{E})$, which serves as the prior knowledge for constructing MPR tasks for a certain user.
We can generate multiple specific graphs $\tilde{G}$ based on the meta graph, thereby constructing different users.


\textbf{Step 3: Collect User Statements.}
After that, we rewrite each non-value-oriented edge in graph $\tilde{G}$ to obtain textual statements for users.
Specifically, we instantiate the information from edges and their connected nodes into prompts, and leverage LLMs to generate user statements $\mathcal{S} = \{s_1, s_2, ..., s_{|\mathcal{S}|}\}$, which serve as single-hop personalized information for users.


\textbf{Step 4: Sample Reasoning Paths.}
We sample a reasoning path $T = \left[ t_1, t_2, ..., t_k \right]$ on graph $\tilde{G}$ as the reference for constructing $k$-hop $(k \ge 2)$ questions.
Specifically, we randomly select an initial edge $t_1 = \left< p_0, p_1 \right> \in \tilde{E}$ as the first edge, and use its endpoint $p_1$ as the starting node to sample the next edge $t_2 = \left< p_1, p_2 \right> \in \tilde{E}$ from its adjacent edges.
To prevent ambiguity in attribute-oriented edges (\textit{i.e.,} multiple entities possessing the same attribute value), we implement the disambiguation mechanism.
When such an ambiguous edge $t_{i} = \left< p_{i-1}, p_i \right>$ occurs, we iteratively sample other attribute-oriented edges from the target entity $p_{i-1}$ until no ambiguity exists.
We iterate for $k$ steps to construct a $k$-hop reasoning path $T$ and ensure that personalized edges are major among the edges. 


\textbf{Step 5: Derive Questions and Answers.}
Based on the reasoning path $T = \left[ t_1, t_2, ..., t_k \right]$, we get the user statement $m_i$ corresponding to the edge $t_i$, obtaining the references of user statements $M = \left[ m_1, m_2, ..., m_k \right] (m_i \in \mathcal{S}, 1 \le i \le k)$.
Then, we instruct LLMs to rewrite them as a multi-hop reasoning question $q$, whose answer is the endpoint of $T$.
Finally, we construct the user statements $\mathcal{S}$ as the training or retrieval corpus, and formulate quadruples $\mathcal{D} = \{(T_i, M_i, q_i, a_i)\}_{i=1}^m$ to represent $m$ testing tasks.
To enhance the diversity of users for evaluation, we sample $n$ training sets and testing sets (\textit{i.e.,} sub-datasets) as final dataset $\mathcal{H} = \{(\mathcal{S}_i, \mathcal{D}_i)\}_{i=1}^n$.



\subsubsection{MPR Dataset Statistics.}
The MPR dataset contains a total of 10,800 multi-hop reasoning QA tasks, covering questions ranging from 2-hop to 10-hop to explicitly represent different reasoning difficulties.
For each user, we construct over 13,000 user statements as personalized information, and build 1,000 QA tasks for each hop count ranging from 2 to 9.
In addition, the MPR dataset contains 12 different specific graphs $\tilde{G}$.
More detailed statistics of the MPR dataset are available in \textbf{Table~\ref{tab:dataset_statistics}}.

\begin{table}[t]
	\centering
	\caption{The statistics of the MPR datsest.}
	\vspace{-0.2cm}
	\resizebox{\linewidth}{!}{
	\begin{tabular}{cccc}
		\hline
		\hline
		\textbf{Statistic} & \textbf{Number} & \textbf{Statistic} & \textbf{Number} \bigstrut\\
		\hline
		Statement (Total) & 157,166 & QA (Total) & 108,000 \bigstrut[t]\\
		Statement (Piece/User) & 13,097$\pm$6.34 & QA (Each User) & 9,000 \\
		Statement (Word/User) & 104,376$\pm$846 & QA (Hop) & 2$\sim$10 \\
		Graph Node (Each User) & 5,902$\pm$43.1 & Entity Type & 108 \\
		Graph Edge (Each User) & 89,586$\pm$59,259 & Relation Type & 34 \bigstrut[b]\\
		\hline
		\hline
	\end{tabular}}
	\label{tab:dataset_statistics}%
	\vspace{-0.4cm}
\end{table}%

\subsection{Experimental Settings}
\subsubsection{Overview.}
The evaluation pipeline of MPR tasks is demonstrated in \textbf{Figure~\ref{fig:pipeline}(b)}.
First, for each sub-dataset $(\mathcal{S}_i, \mathcal{D}_i), 1 \le i \le n$, we provide the collection of user statements $S_i$ to the memory model of agents, and allow it to perform preprocessing, such as building indices or conducting fine-tuning.
Then, we conduct the evaluation on questions in $\mathcal{D}_i$.
Specifically, for each testing point $(T, M, q, a) \in \mathcal{D}_i$, we only present the question $q$ to agents to obtain a predicted answer $\hat{a}$ by multi-hop reasoning according to $\mathcal{S}_i$. 
At this point, the reasoning path $T$, references $M$, and answer $a$ are considered as ground-truths that are invisible to the agents.
Finally, we compare the predicted answer $\hat{a}$ with the ground-truth answer $a$ after normalization, and calculate the average accuracy (ACC) using Exact Match (EM)~\cite{rajpurkar2016squad} as follows:
\setlength\abovedisplayskip{3pt}
\setlength\belowdisplayskip{3pt}
\begin{gather*}
	\text{ACC} = \frac{1}{|\mathcal{H}|} \sum_{(\mathcal{S}_i, \mathcal{D}_i) \in \mathcal{H}} \frac{1}{|\mathcal{D}_i|} \sum_{(T_j, M_j, q_j, a_j) \in \mathcal{D}_i} \text{EM} \left[ f(q_j; \mathcal{S}_j), a_j \right],
\end{gather*}
where $f$ represents the evaluated model (\textit{i.e.,} LLM-based agent).
In order to enable evaluating more aspects at a finer granularity, we select 2,700 questions (uniformly across 2-hop to 10-hop) from the entire dataset as a sub-dataset to reduce computational overhead.
Our experiments are consistently conducted on a server equipped with 8 NVIDIA A800-SXM-80G GPUs.
To control for confounding factors, we use Qwen2.5-7B~\cite{qwen2.5} as our base model for all experiments except exploring different model sizes.


\subsubsection{Memory Mechanisms.}
Our study primarily explores the performance of explicit memory and implicit memory in MPR tasks. Therefore, we respectively utilize explicit, implicit, and hybrid memory methods to store and utilize user statements, as illustrated in \textbf{Figure~\ref{fig:pipeline}(c)}.
For explicit memory, we store user statements in textual form and build indices during the training phase, and retrieve based on the current query when testing.
We commonly set the retrieval count to $k=20$ (\textit{i.e.,} top-20 statements), adopt BM25s~\cite{bm25s} for all sparse retrievals, and utilize e5-base-v2~\cite{wang2022text} for dense retrievals.
For implicit memory, we fine-tune the base model according to user statements during training, and perform inference using the fine-tuned model.
Specifically, we adopt LoRA~\cite{hu2022lora} to conduct fine-tuning and set the LoRA rank to 8 and alpha to 32.
We tune the training epochs for all methods ranging from 1 to 10.
For hybrid memory, we store and build indices while fine-tuning the model during the training phase, and utilize the trained model with retrieved information for testing.
The detailed baselines will be described and discussed in the later sections.

\subsubsection{Reasoning Structures.}
Since MPR tasks require multi-hop reasoning, we implement several different reasoning structures to better explore the performance of memory under different reasoning strategies, as shown in \textbf{Figure~\ref{fig:pipeline}(d)} and described as follows:\\
$\bullet$ \textbf{Naive Reasoning (NR):} the vanilla reasoning method that obtains the predicted answer through only a single-step LLM inference, without using multi-hop reasoning structures.\\
$\bullet$ \textbf{Sequential Reasoning (SR):} the chain-like reasoning where each inference is based on the results from the last step, forming a sequence of multiple LLM inferences, implemented by CoT~\cite{wei2022chain}.\\
$\bullet$ \textbf{Multi-path Reasoning (MR):} the extension of sequential reasoning that maintains multiple chains and chooses one for extended reasoning at each inference step, implemented by ToT~\cite{yao2023tree}.\\
$\bullet$ \textbf{Decomposition Reasoning (DR):} the divide-and-conquer structure~\cite{wang2023plan} that decomposes a task into multiple sub-tasks, then processes them and combines their results to infer the final answer.


\begin{figure}[t]
	\centering
	\begin{subfigure}{\linewidth}
		\includegraphics[width=1.0\textwidth]{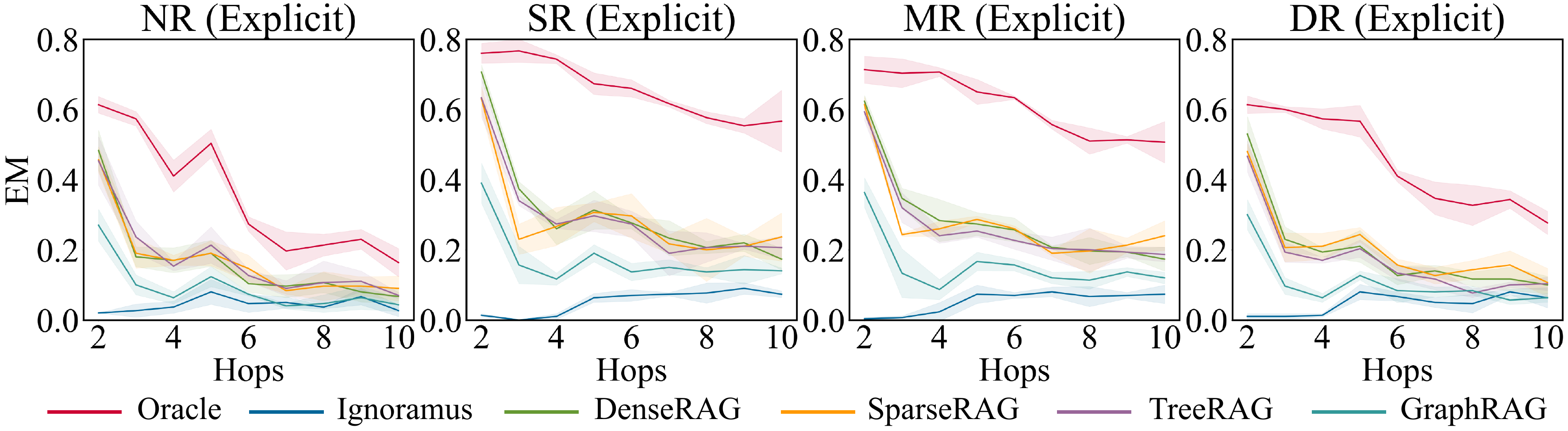}
	\end{subfigure}
	\vspace{-0.65cm}
	\caption{Overall performances of explicit memory, with mean values (line) and standard deviation values (shading).}
	\label{fig:rag_major}
	\vspace{-0.6cm}
\end{figure}

For different reasoning structures, we carefully design the reasoning prompts while ensuring consistent basic content.
More details are available in \textbf{Appendix~\ref{appendix:reasoning_structures}}.
We set the number of multi-hop reasoning steps to 5, except for experiments exploring the impact of reasoning steps.
For the Multi-path Reasoning structure, we set the number of augmented branches to 2 at each step.

\section{Explicit Memory on MPR Task}
\subsection{Overview}
First of all, we explore the performance of explicit memory for LLM-based agents on MPR tasks.
To ensure fair evaluation, we adopt the general RAG pipeline~\cite{fan2024survey}, which retrieves relevant statements based on the current query (or reasoning state) and integrates them into the prompt to facilitate LLM inference.
The baselines of RAG methods for explicit memory are as follows:\\
$\bullet$ \textbf{SparseRAG:} an unstructured RAG method that represents statements and queries as sparse vectors based on  TF-IDF~\cite{sparck1972statistical}, leveraging exact keyword matching and statistical term importance.\\
$\bullet$ \textbf{DenseRAG:} an unstructured RAG method
that utilizes neural embedding models~\cite{wang2022text} to encode statements and queries into dense vectors, employing cosine similarity to get contextual relevance.\\
$\bullet$ \textbf{TreeRAG:} a tree-structured RAG method that represents each statement as a leaf node and considers parent nodes as summaries of their child nodes hierarchically, implemented as MemTree~\cite{rezazadeh2024isolated}.\\
$\bullet$ \textbf{GraphRAG:} a graph-structured RAG method that extracts entities and relations from each statement and constructs a knowledge graph~\cite{peng2024graph}.
During retrieval, it calculates the semantic relevance between the query and entities/relations.

Besides the methods mentioned above, we establish two special baselines for more comprehensive studies as follows:\\
$\bullet$ \textbf{Ignoramus:} an ablation baseline for comparison, which conducts inference without using any user statements.\\
$\bullet$ \textbf{Oracle:} an upper-bound baseline that utilizes additional golden references for inference, eliminating retrieval errors.

In the following sections, we evaluate all these explicit memory baselines across all reasoning structures. We analyze the experimental results and draw conclusions.

\begin{figure}[tb]
	\centering
        \vspace{-0.05cm}
	\begin{subfigure}{\linewidth}
		\includegraphics[width=1.0\textwidth]{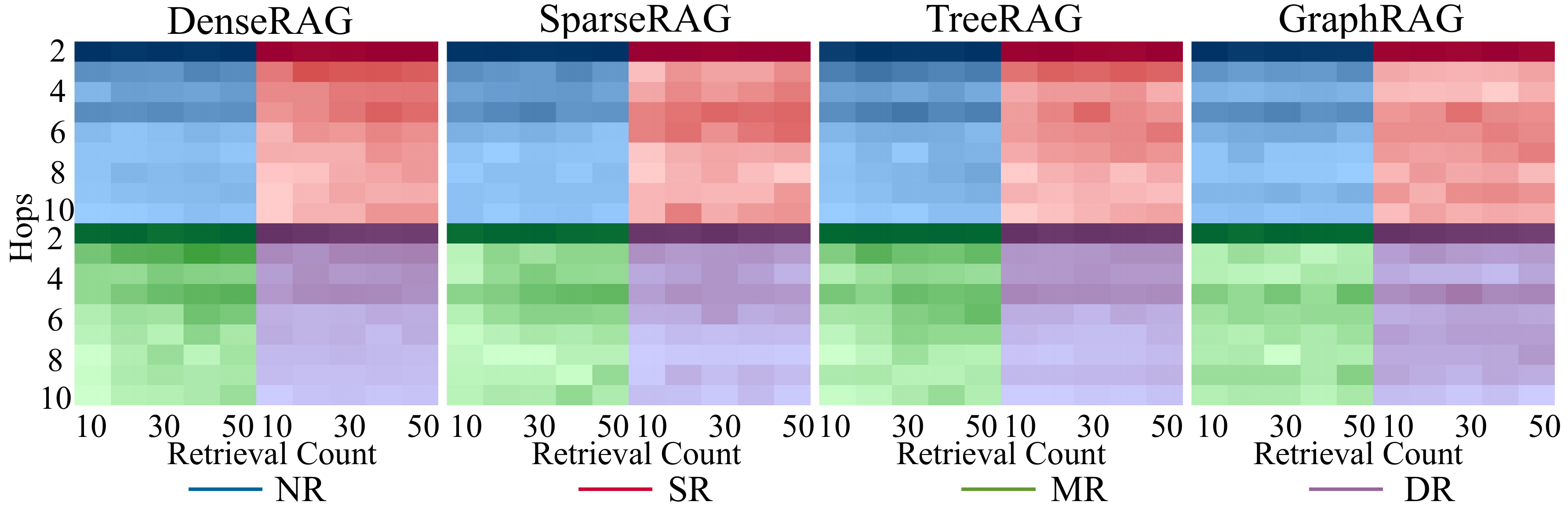}
	\end{subfigure}
	\vspace{-0.75cm}
	\caption{Performance of various retrieved statement counts. Darker colors indicate higher accuracy in MPR tasks.}
	\label{fig:rag_topk}
	\vspace{-0.55cm}
\end{figure}

\subsection{Overall Performances}
The overall performances are presented in \textbf{Figure~\ref{fig:rag_major}}.
From the results, we find that reasoning structures significantly affect task performance in explicit memory.
The performance of SR and MR is significantly higher than DR and NR, with an overall improvement of 10\% to 20\%.
Besides, NR is weaker than other multi-hop structures, especially on long-hop questions.
It indicates that the test-time scaling~\cite{zhang2025survey_tts} is effective, and multi-hop reasoning structures are important for complex problems.
Additionally, DR relies on the initial question for task decomposition, which may exhibit certain myopia that affects further reasoning.

From the perspective of memory baselines, we find DenseRAG performs best on short-hop questions, while SparseRAG performs best on long-hop questions.
This may be because DenseRAG can better locate required information based on semantic granularity, while SparseRAG can retrieve broader information based on word granularity.
Besides, TreeRAG performs comparably to the above two methods. To our surprise, GraphRAG shows poor performance in our settings, which may be affected by similar entities during graph construction, thereby introducing much noise.

In addition, as the question hop increases, the overall accuracy shows a declining trend as expected.
For example, the accuracy of DenseRAG on SR and MR gradually decreases from over 60\% on 2-hop questions to around 20\% on 10-hop questions.
Besides, we find SR and MR can mitigate the decline on Oracle, but other RAG methods still exhibit rapid degradation.
As the difficulty of questions increases, the requirement of multi-hop reasoning also boosts, so SR and MR greatly improve the performance of long-hop questions on Oracle.
However, other baselines may lack a global perspective due to partial retrieval, thereby reducing planning abilities.

\begin{figure}[t]
	\centering
	\begin{subfigure}{\linewidth}
		\includegraphics[width=1.0\textwidth]{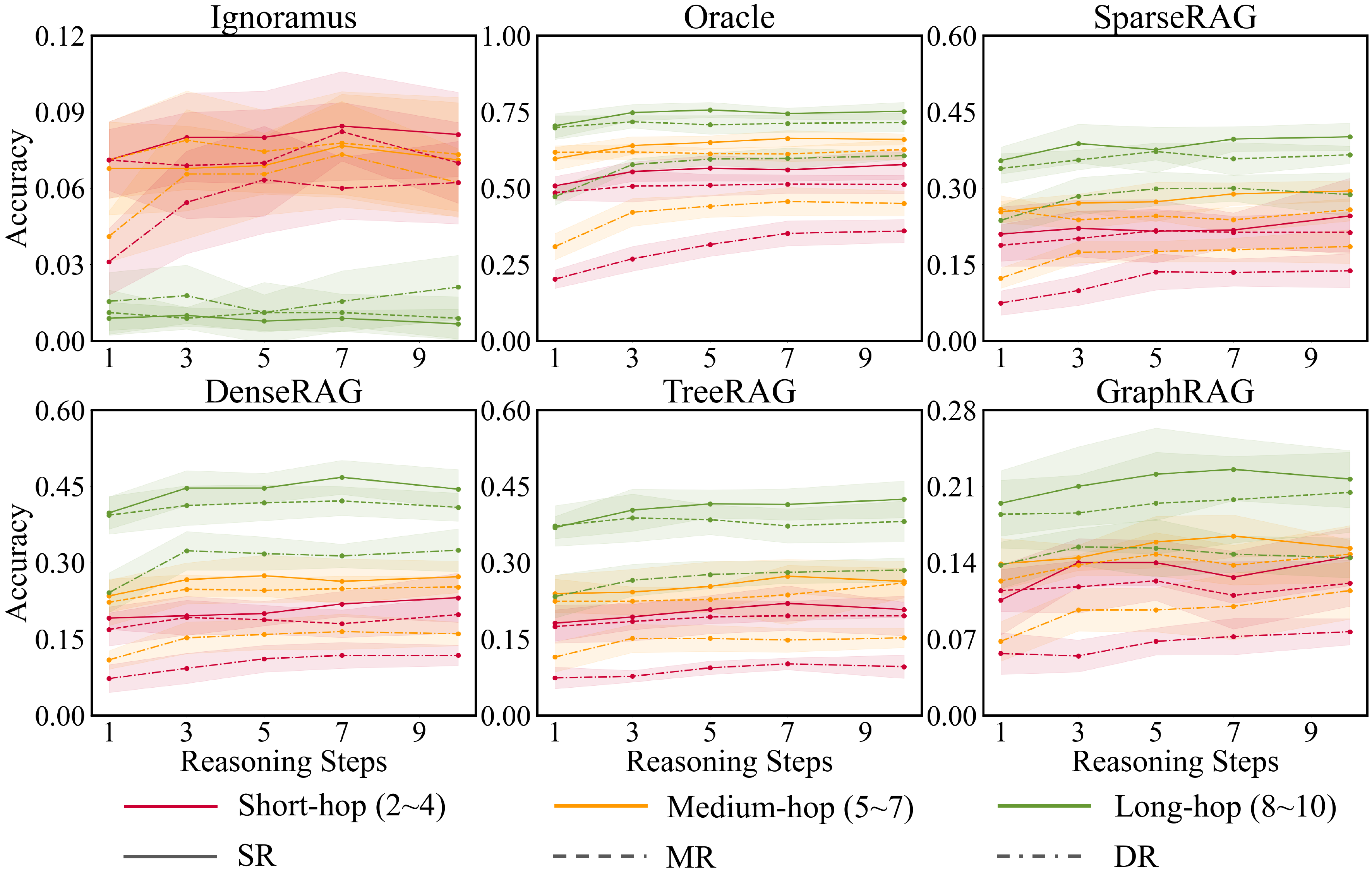}
	\end{subfigure}
	
	\vspace{-0.4cm}
	\caption{Performance of various reasoning steps, with mean values (line) and standard deviation values (shading).}
	\label{fig:rag_restep_final}
	\vspace{-0.65cm}
\end{figure}

\subsection{Impact of Retrieved Statement Counts}
For explicit memory, the number of retrieved information is a critical factor.
Too few retrievals may result in information loss, while too many can lead to noise and extra cost.
Therefore, we further explore the impact of retrieved statement counts.
Specifically, we construct heatmaps to analyze four RAG baselines, and use different colors to represent reasoning structures, as presented in \textbf{Figure~\ref{fig:rag_topk}}. 

According to the results, we find that in short-hop questions, the accuracy typically improves as $k$ increases.
We speculate short-hop questions require recalling more comprehensive information to facilitate short-range reasoning, so a larger $k$ can provide more sufficient information (\textit{i.e.,} high recall).
However, in long-hop questions, there exists a peak of accuracy at an intermediate $k$ value.
This may be because long-range reasoning is more susceptible to noise, and too large $k$ values may not be applicable (\textit{i.e.,} low precision).

\subsection{Impact of Reasoning Steps}
For reasoning tasks, the number of reasoning steps can be a critical bottleneck.
Increasing the number of reasoning steps typically leads to higher accuracy, but also results in linear growth of reasoning costs.
To further study the impact on model performance, we set different numbers of reasoning steps and conduct experiments across various memory baselines and reasoning structures.
The results are presented in \textbf{Figure~\ref{fig:rag_restep_final}}.

From the results, we find that model performances improve as the number of reasoning steps increases, but the gains beyond 3 steps are not significant.
Furthermore, we observe that DR exhibits greater sensitivity compared to the other two reasoning structures.
This may be because extending the reasoning chain can potentially mitigate the limitation of initial question decomposition.
Additionally, long-hop problems are more affected compared to short-hop and medium-hop questions, which is intuitive as long-hop problems require more reasoning steps.
Finally, Oracle is more affected than other RAG baselines, possibly because the model with golden references benefits more from increased reasoning steps.

\subsection{Impact of Backbone Sizes}
\begin{figure}[t]
	\centering
        \vspace{0.05cm}
	\begin{subfigure}{\linewidth}
		\includegraphics[width=1.0\textwidth]{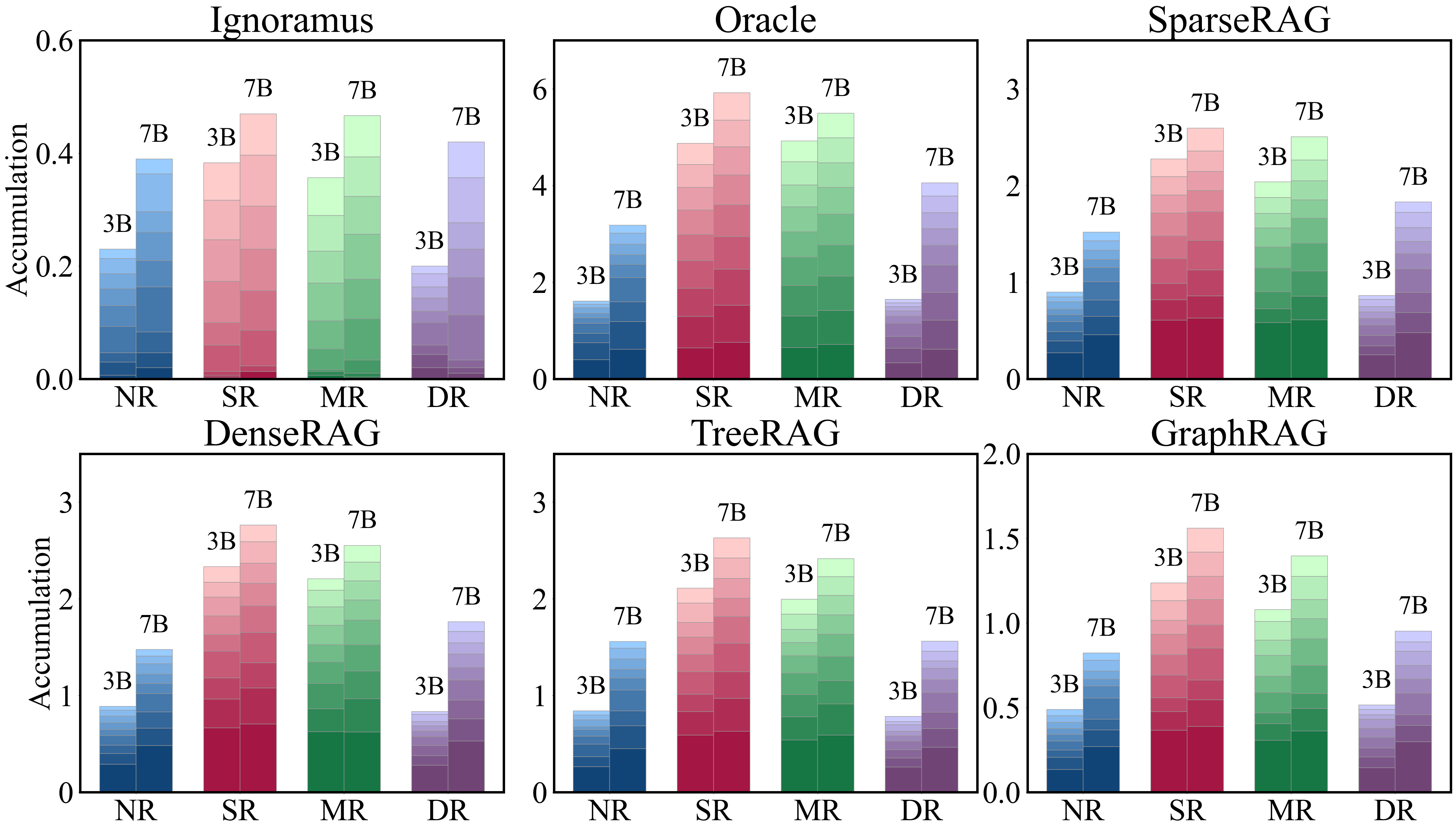}
	\end{subfigure}
	
	\vspace{-0.3cm}
	\caption{Performance of various backbone sizes. In each bar, the cells indicate increasing hops from bottom to top.}
	\label{fig:rag_backbone}
	\vspace{-0.6cm}
\end{figure}

Explicit memory primarily relies on the ICL capability of LLMs, and different backbone sizes exhibit varying abilities for context understanding and reasoning.
Therefore, we further study the performance of various baselines and reasoning structures under different sizes of backbones.
We present the results in \textbf{Figure~\ref{fig:rag_backbone}}.

We observe that the 3B model shows significant degradation compared to the 7B model across all scenarios.
Among reasoning structures, the performance decline of NR and DR is significantly higher than the others, indicating that sequential reasoning demonstrates higher robustness.
Additionally, different RAG baselines exhibit consistent performance degradation, as retrieval and generation are relatively independent processes under our setting.

\subsection{Model Efficiency}
\begin{figure}[t]
	\centering
	\vspace{-0.05cm}
	\begin{subfigure}{\linewidth}
		\includegraphics[width=1.0\textwidth]{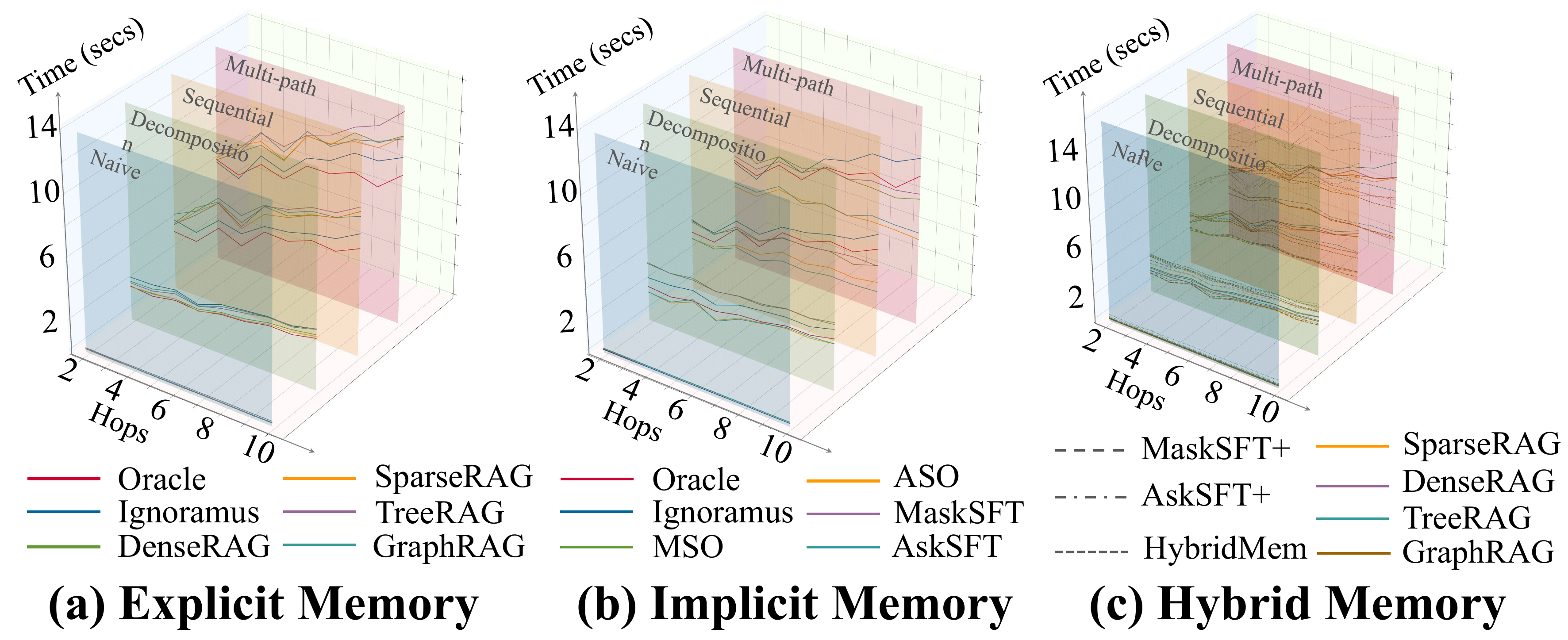}
	\end{subfigure}

	\vspace{-0.3cm}
	\caption{Efficiency of different memory mechanisms.}
	\label{fig:exp_efficiency}
	\vspace{-0.6cm}
\end{figure}

Efficiency is critical for LLM-based agents, affecting user satisfaction and computational overhead.
Efficient memory approaches contribute to improving the response speed of LLM-based agents, especially in online environments.
Therefore, we further evaluate the efficiency of various memory baselines and reasoning structures.
Specifically, we calculate the average task completion time under different question hops. The results are shown in \textbf{Figure~\ref{fig:exp_efficiency}(a)}.

The results indicate that multi-hop reasoning structures demonstrate a substantial increase in time consumption compared to single-step reasoning, because additional reasoning steps bring extra inference time.
Moreover, the RAG baselines exhibit higher time overhead than both the Oracle and Ignoramus, and TreeRAG exhibits significantly higher overhead than the other methods.
This is because TreeRAG constructs extensive summarization nodes for retrieval.
Finally, the more hops a question has, the longer the time consumption, as a higher number of hops implies greater question complexity, leading to more reasoning tokens to reach the answer.

\section{Implicit Memory on MPR Task}
\subsection{Overview}
We further explore the performance of implicit memory in LLM-based agents on MPR tasks.
During the training phase, we transform user statements into input-output tuples and conduct fine-tuning based on LoRA.
During the testing phase, we do not employ retrieval strategies, but perform inference directly.
According to the user statement transformation, we have the following baselines:\\
$\bullet$ \textbf{MaskSFT:} randomly mask entities or relations of user statements as input, and use the masked information as output to conduct instruction fine-tuning, inspired by \citet{devlin2019bert}.\\
$\bullet$ \textbf{AskSFT:} rewrite user statements into QA pairs, and use them as input and output, respectively for instruction fine-tuning.

To better compare the effects of implicit memory, we also add two special baselines, MaskSFT + Oracle (MSO) and AskSFT + Oracle (ASO), which incorporate golden references during inference.

\subsection{Overall Performances}
\begin{figure}[t]
	\centering
	\begin{subfigure}{\linewidth}
		\includegraphics[width=1.0\textwidth]{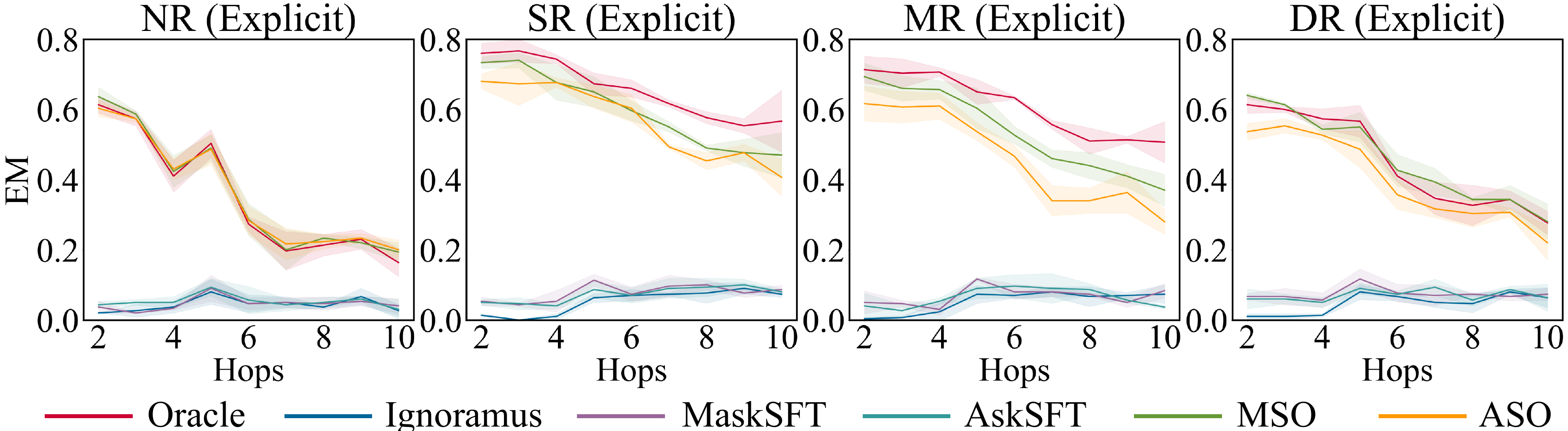}
	\end{subfigure}
	\vspace{-0.6cm}
	\caption{Overall performances of implicit memory.}
	\label{fig:sft_major}
	\vspace{-0.6cm}
\end{figure}

The results of overall performances are presented in \textbf{Figure~\ref{fig:sft_major}}.
We find that across all reasoning structures, using implicit memory alone achieves poor performance, indicating that SFT cannot effectively handle large-scale detailed information.
Besides, there exists consistent degradation on SR and MR by adding implicit memory on Oracle baselines, possibly because SFT may decline the reasoning capability of LLMs.
We also find that on DR, ASO shows certain improvement over Oracle, especially on long-hop problems, which may be due to the enhancement of task decomposition capability.

\subsection{Impact of Training Steps}
The number of training epochs is significant for SFT, as too many epochs can cause overfitting, while too few epochs may lead to underfitting.
Therefore, we further explore the model performance under different training epochs.
Due to the page limitation, we put the results in \textbf{Appendix~\ref{appendix:sft_steps}}.
The results indicate that implicit memory alone cannot achieve great performance even after training for more epochs.
Additionally, the performance of MSO and ASO significantly declines as training steps increase, possibly because more training steps lead to greater reasoning capability degradation.

\subsection{Impact of Reasoning Steps}
Similar to explicit memory, we also explore the impact of reasoning steps on implicit memory, with results shown in \textbf{Appendix~\ref{appendix:sft_resteps}} due to the page limitation.
We find that even with increased reasoning steps, implicit memory cannot achieve reasonable results under our setting. For MSO and ASO, their results exhibit similar patterns to explicit memory as analyzed in the previous section.

\subsection{Impact of Base Model Sizes}
We also study the impact of model size on implicit memory, with results presented in \textbf{Appendix~\ref{appendix:sft_backbone}} due to the page limitation.
Across various reasoning structures and memory baselines, the 3B model shows significant degradation in accuracy.
Among different reasoning structures, NR and DR exhibit the most pronounced performance decline, significantly higher than the other two reasoning structures, with DR declining by approximately 80\%.

\subsection{Model Efficiency}
We conduct an efficiency analysis of implicit memory approaches, with results illustrated in \textbf{Figure~\ref{fig:exp_efficiency}(b)}.
Beyond the patterns consistent with explicit memory findings, we find that implicit memory does not require retrieval overhead and extracts tokens to contain retrieved content. It suggests that implicit memory approaches offer considerable potential for enhancing inference efficiency.

\begin{table*}[t]
	\centering
	\caption{Overall performances of different hybrid memory methods. The sign +X represents direct combination of various explicit memory methods, and the bold values indicate the best performances.}
	\vspace{-0.4cm}
	\resizebox{\linewidth}{!}{
		\begin{tabular}{cccccccccccc}
			\hline
			\hline
			& \multicolumn{6}{c}{\textbf{Naive Reasoning}}  & \multicolumn{5}{c}{\textbf{Sequential Reasoning}} \bigstrut\\
			\hline
			\textbf{Hops} & \textbf{Methods} & \textbf{Oracle} & \textbf{DenseRAG} & \textbf{SparseRAG} & \textbf{TreeRAG} & \textbf{GraphRAG} & \textbf{Oracle} & \textbf{DenseRAG} & \textbf{SparseRAG} & \textbf{TreeRAG} & \textbf{GraphRAG} \bigstrut\\
			\hline
			\multicolumn{1}{c}{\multirow{4}[2]{*}{Short\newline{}(2$\sim$6)}} & X     & 0.428  & \textbf{0.206} & 0.204  & 0.212  & 0.112  & \textbf{0.703} & 0.324  & 0.361  & 0.334  & 0.190  \bigstrut[t]\\
			& MS+X  & \textbf{0.437} & 0.204  & 0.207  & 0.202  & 0.106  & 0.658  & 0.289  & 0.319  & 0.307  & 0.174  \\
			& AS+X  & 0.432  & 0.212  & \textbf{0.212} & \textbf{0.213} & \textbf{0.117} & 0.627  & 0.263  & 0.287  & 0.269  & 0.164  \\
			& \cellcolor[rgb]{ .867,  .922,  .969} HybridMem & \cellcolor[rgb]{ .867,  .922,  .969} 0.436  & \cellcolor[rgb]{ .867,  .922,  .969} 0.197  & \cellcolor[rgb]{ .867,  .922,  .969} 0.202  & \cellcolor[rgb]{ .867,  .922,  .969} 0.207  & \cellcolor[rgb]{ .867,  .922,  .969} 0.114  & \cellcolor[rgb]{ .867,  .922,  .969} 0.702  & \cellcolor[rgb]{ .867,  .922,  .969} \textbf{0.326} & \cellcolor[rgb]{ .867,  .922,  .969} \textbf{0.368} & \cellcolor[rgb]{ .867,  .922,  .969} \textbf{0.336} & \cellcolor[rgb]{ .867,  .922,  .969} \textbf{0.192} \bigstrut[b]\\
			\hline
			\multicolumn{1}{c}{\multirow{4}[2]{*}{Long\newline{}(7$\sim$10)}} & X     & 0.202  & 0.094  & 0.084  & \textbf{0.096} & 0.051  & \textbf{0.566} & 0.216  & 0.200  & 0.208  & \textbf{0.140} \bigstrut[t]\\
			& MaskSFT+X & 0.216  & 0.086  & 0.081  & 0.084  & 0.050  & 0.479  & 0.190  & 0.190  & 0.174  & 0.121  \\
			& AskSFT+X & \textbf{0.219} & \textbf{0.097} & 0.089  & \textbf{0.096} & \textbf{0.058} & 0.446  & 0.168  & 0.170  & 0.134  & 0.112  \\
			& \cellcolor[rgb]{ .867,  .922,  .969} HybridMem & \cellcolor[rgb]{ .867,  .922,  .969} 0.208  & \cellcolor[rgb]{ .867,  .922,  .969} 0.094  & \cellcolor[rgb]{ .867,  .922,  .969} \textbf{0.090} & \cellcolor[rgb]{ .867,  .922,  .969} 0.093  & \cellcolor[rgb]{ .867,  .922,  .969} \textbf{0.058} & \cellcolor[rgb]{ .867,  .922,  .969} 0.563  & \cellcolor[rgb]{ .867,  .922,  .969} \textbf{0.223} & \cellcolor[rgb]{ .867,  .922,  .969} \textbf{0.232} & \cellcolor[rgb]{ .867,  .922,  .969} \textbf{0.209} & \cellcolor[rgb]{ .867,  .922,  .969} 0.139  \bigstrut[b]\\
			\hline
			\hline
			& \multicolumn{6}{c}{\textbf{Multi-path Reasoning}} & \multicolumn{5}{c}{\textbf{Decomposition Reasoning}} \bigstrut\\
			\hline
			\textbf{Hops} & \textbf{Methods} & \textbf{Oracle} & \textbf{DenseRAG} & \textbf{SparseRAG} & \textbf{TreeRAG} & \textbf{GraphRAG} & \textbf{Oracle} & \textbf{DenseRAG} & \textbf{SparseRAG} & \textbf{TreeRAG} & \textbf{GraphRAG} \bigstrut\\
			\hline
			\multicolumn{1}{c}{\multirow{4}[2]{*}{Short\newline{}(2$\sim$6)}} & X     & 0.661  & \textbf{0.309} & \textbf{0.332} & 0.306  & \textbf{0.171} & 0.518  & \textbf{0.237} & 0.238  & \textbf{0.214} & 0.125  \bigstrut[t]\\
			& MS+X  & 0.600  & 0.263  & 0.286  & 0.276  & 0.153  & 0.528  & 0.228  & 0.234  & 0.211  & \textbf{0.132} \\
			& AS+X  & 0.529  & 0.202  & 0.195  & 0.208  & 0.121  & 0.463  & 0.222  & 0.219  & 0.201  & 0.124  \\
			& \cellcolor[rgb]{ .867,  .922,  .969} HybridMem & \cellcolor[rgb]{ .867,  .922,  .969} \textbf{0.672} & \cellcolor[rgb]{ .867,  .922,  .969} 0.304  & \cellcolor[rgb]{ .867,  .922,  .969} 0.326  & \cellcolor[rgb]{ .867,  .922,  .969} \textbf{0.310} & \cellcolor[rgb]{ .867,  .922,  .969} 0.166  & \cellcolor[rgb]{ .867,  .922,  .969} \textbf{0.533} & \cellcolor[rgb]{ .867,  .922,  .969} 0.234  & \cellcolor[rgb]{ .867,  .922,  .969} \textbf{0.239} & \cellcolor[rgb]{ .867,  .922,  .969} 0.210  & \cellcolor[rgb]{ .867,  .922,  .969} 0.122  \bigstrut[b]\\
			\hline
			\multicolumn{1}{c}{\multirow{4}[2]{*}{Long\newline{}(7$\sim$10)}} & X     & 0.510  & \textbf{0.217} & 0.188  & 0.193  & \textbf{0.123} & 0.316  & 0.136  & 0.111  & 0.093  & 0.068  \bigstrut[t]\\
			& MS+X  & 0.407  & 0.159  & 0.166  & 0.156  & 0.100  & 0.322  & 0.128  & 0.099  & 0.104  & \textbf{0.073} \\
			& AS+X  & 0.328  & 0.151  & 0.123  & 0.116  & 0.086  & 0.277  & 0.119  & 0.103  & \textbf{0.099} & 0.068  \\
			& \cellcolor[rgb]{ .867,  .922,  .969} HybridMem & \cellcolor[rgb]{ .867,  .922,  .969} \textbf{0.518} & \cellcolor[rgb]{ .867,  .922,  .969} 0.209  & \cellcolor[rgb]{ .867,  .922,  .969} \textbf{0.194} & \cellcolor[rgb]{ .867,  .922,  .969} \textbf{0.206} & \cellcolor[rgb]{ .867,  .922,  .969} \textbf{0.123} & \cellcolor[rgb]{ .867,  .922,  .969} \textbf{0.321} & \cellcolor[rgb]{ .867,  .922,  .969} \textbf{0.144} & \cellcolor[rgb]{ .867,  .922,  .969} \textbf{0.113} & \cellcolor[rgb]{ .867,  .922,  .969} 0.089  & \cellcolor[rgb]{ .867,  .922,  .969} 0.069  \bigstrut[b]\\
			\hline
			\hline
	\end{tabular}}
	\label{tab:hybrid_major}
	\vspace{-0.3cm}
\end{table*}%

\section{Hybrid Memory on MPR Task}
\subsection{Overview}
Our experiments reveal that relying solely on implicit memory is insufficient for MPR tasks.
However, implicit memory demonstrates potential for enhancing reasoning capabilities.
Therefore, we further combine implicit memory with explicit memory to form hybrid memory for additional experiments.
A straightforward approach involves using SFT models as backbones with retrieved statements for reasoning, resulting in 8 hybrid models.

\subsection{BlockSFT Method}
Although directly combining SFT and RAG is easy to implement, it has several limitations.
First, SFT struggles to capture a large number of details and may encounter conflicts in the constructed instructions.
Besides, training on the entire corpus can degrade reasoning performance.
Finally, the inference process is not query-related, resulting in limited enhancement of explicit memory.

To address these limitations, we propose a novel hybrid method called HybridMem.
In MPR tasks, user statements typically exhibit contextual relationships, forming multiple local clusters.
Therefore, we divide the entire collection of user statements into multiple clusters and conduct SFT independently on each cluster to obtain multiple LoRA adapters while constructing corresponding indices.
During inference, we identify the most relevant LoRA adapter based on the current retrieved statements and conduct inference using the selected adapter.
Specifically, we employ K-means~\cite{lloyd1982least} for clustering, adopt a voting aggregation strategy to determine the adapter most relevant to the retrieval results, and use the same dense retrieval method as explicit memory to construct indices.

\subsection{Overall Performance}
The overall performance results are presented in \textbf{Table~\ref{tab:hybrid_major}}.
Our method achieves the best overall performance on multi-hop reasoning structures, with particularly notable improvements on long-hop questions. This demonstrates the effectiveness of our proposed approach.
Additionally, we observe that implicit memory can degrade model performance, especially for AS+X under multi-hop reasoning scenarios. This degradation likely results from a trade-off between the implicit memory's command of overall knowledge and the reasoning degradation introduced during training. Finally, we find that HybridMem consistently improves performance across multiple RAG methods and also enhances the Oracle baseline.

\begin{figure}[t]
	\centering
	\begin{subfigure}{\linewidth}
		\includegraphics[width=1.0\textwidth]{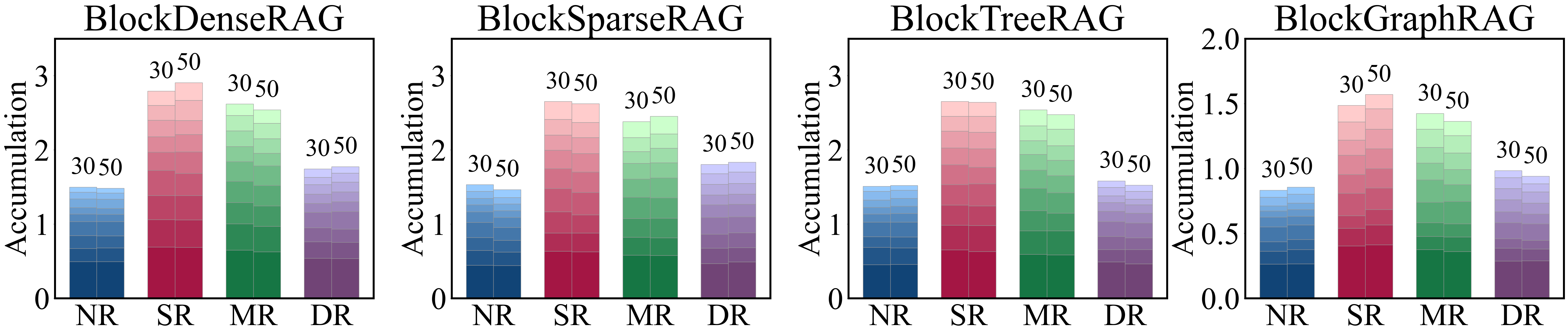}
	\end{subfigure}
	\vspace{-0.75cm}
	\caption{Performance of various cluster counts. In each bar, the cells indicates increasing hops from bottom to top.}
	\label{fig:HybridMem_block_num}
	\vspace{-0.6cm}
\end{figure}

\subsection{Impact of Cluster Counts}
The number of clusters used to divide the entire statement collection is a crucial factor. Fewer clusters require each LoRA adapter to carry more information, while more clusters incur greater training overhead. Therefore, we evaluate the performance across different cluster numbers, with results shown in \textbf{Figure~\ref{fig:HybridMem_block_num}}.
We find that the performance difference between 30 blocks and 50 blocks is minimal, though the 50-block configuration shows slight improvements. This indicates that our method exhibits good robustness to block counts.

\subsection{Impact of Training Epochs}
\begin{figure}[t]
	\centering
	\begin{subfigure}{\linewidth}
		\includegraphics[width=1.0\textwidth]{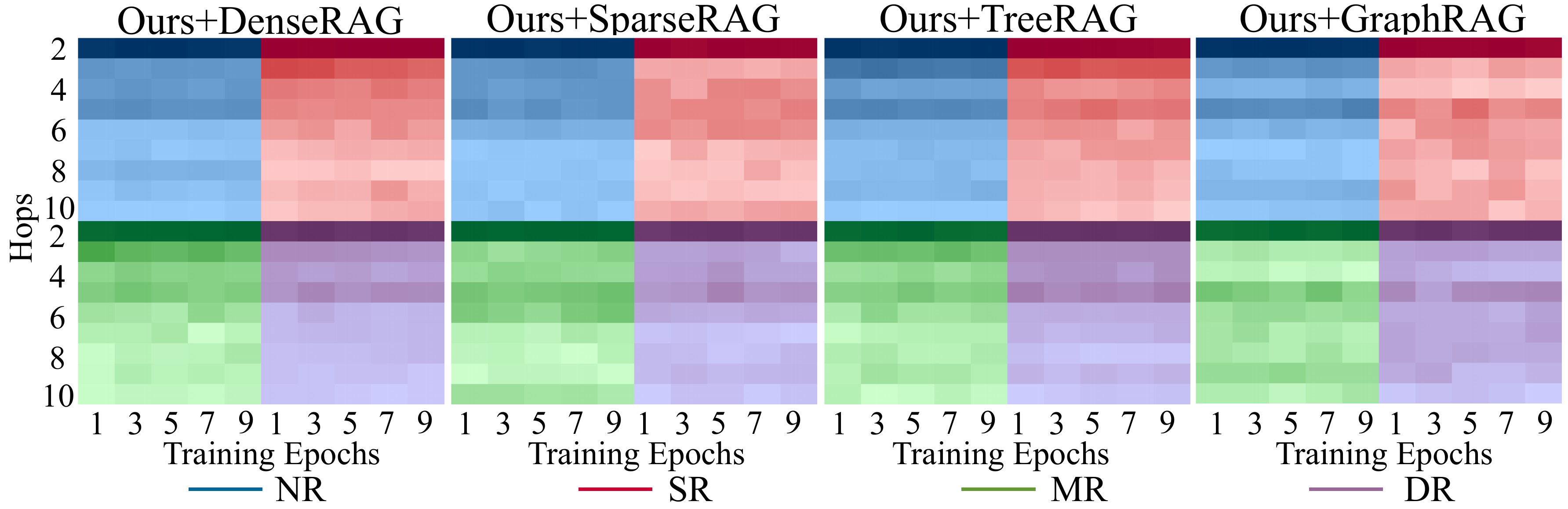}
	\end{subfigure}
	
	\vspace{-0.3cm}
	\caption{Performance of training epochs. Darker colors indicate higher accuracy in MPR tasks.}
	\label{fig:hybrid_steps}
	\vspace{-0.6cm}
\end{figure}

Additionally, we explore the impact of different training epochs on our method, with results shown in \textbf{Figure~\ref{fig:hybrid_steps}}.
The results show that model performance initially improves before decreasing as the number of training epochs increases.
The model achieves optimal performance at epoch 7, particularly for long-hop questions.
Furthermore, for different reasoning structures, we notice that the number of training epochs significantly affects SR and MR performance, but shows minimal impact on NR and DR.


\subsection{Model Efficiency}
In addition, we evaluate the efficiency of various hybrid models, with results presented in \textbf{Figure~\ref{fig:exp_efficiency}(c)}.
The results show that HybridMem generally requires more time than other approaches, particularly on SR and MR structures.
This is attributed to the additional cost of statement clustering and adapter loading.
Furthermore, HybridMem significantly increases the differences in time requirements across various RAG methods.

\section{Conclusion}
In this paper, we formally define the MPR task and construct a dataset to evaluate different memory approaches.
We conduct experiments with various types of memory on MPR tasks, drawing conclusions and findings.
We propose HybridMem to better solve long-hop tasks. 
In future research, we will explore the adaptive integration of implicit memory and explicit memory, as well as multimodal personalized memory with reasoning strategies.

\clearpage

\bibliographystyle{ACM-Reference-Format}
\bibliography{references}


\begin{thebibliography}{48}


\ifx \showCODEN    \undefined \def \showCODEN     #1{\unskip}     \fi
\ifx \showISBNx    \undefined \def \showISBNx     #1{\unskip}     \fi
\ifx \showISBNxiii \undefined \def \showISBNxiii  #1{\unskip}     \fi
\ifx \showISSN     \undefined \def \showISSN      #1{\unskip}     \fi
\ifx \showLCCN     \undefined \def \showLCCN      #1{\unskip}     \fi
\ifx \shownote     \undefined \def \shownote      #1{#1}          \fi
\ifx \showarticletitle \undefined \def \showarticletitle #1{#1}   \fi
\ifx \showURL      \undefined \def \showURL       {\relax}        \fi
\providecommand\bibfield[2]{#2}
\providecommand\bibinfo[2]{#2}
\providecommand\natexlab[1]{#1}
\providecommand\showeprint[2][]{arXiv:#2}

\bibitem[Chang et~al\mbox{.}(2024)]%
        {chang2024survey}
\bibfield{author}{\bibinfo{person}{Yupeng Chang}, \bibinfo{person}{Xu Wang},
  \bibinfo{person}{Jindong Wang}, \bibinfo{person}{Yuan Wu},
  \bibinfo{person}{Linyi Yang}, \bibinfo{person}{Kaijie Zhu},
  \bibinfo{person}{Hao Chen}, \bibinfo{person}{Xiaoyuan Yi},
  \bibinfo{person}{Cunxiang Wang}, \bibinfo{person}{Yidong Wang},
  {et~al\mbox{.}}} \bibinfo{year}{2024}\natexlab{}.
\newblock \showarticletitle{A survey on evaluation of large language models}.
\newblock \bibinfo{journal}{\emph{ACM transactions on intelligent systems and
  technology}} \bibinfo{volume}{15}, \bibinfo{number}{3}
  (\bibinfo{year}{2024}), \bibinfo{pages}{1--45}.
\newblock


\bibitem[Chen et~al\mbox{.}(2024)]%
        {chen2024large}
\bibfield{author}{\bibinfo{person}{Jin Chen}, \bibinfo{person}{Zheng Liu},
  \bibinfo{person}{Xu Huang}, \bibinfo{person}{Chenwang Wu},
  \bibinfo{person}{Qi Liu}, \bibinfo{person}{Gangwei Jiang},
  \bibinfo{person}{Yuanhao Pu}, \bibinfo{person}{Yuxuan Lei},
  \bibinfo{person}{Xiaolong Chen}, \bibinfo{person}{Xingmei Wang},
  {et~al\mbox{.}}} \bibinfo{year}{2024}\natexlab{}.
\newblock \showarticletitle{When large language models meet personalization:
  Perspectives of challenges and opportunities}.
\newblock \bibinfo{journal}{\emph{World Wide Web}} \bibinfo{volume}{27},
  \bibinfo{number}{4} (\bibinfo{year}{2024}), \bibinfo{pages}{42}.
\newblock


\bibitem[Chen et~al\mbox{.}(2020)]%
        {chen2020hybridqa}
\bibfield{author}{\bibinfo{person}{Wenhu Chen}, \bibinfo{person}{Hanwen Zha},
  \bibinfo{person}{Zhiyu Chen}, \bibinfo{person}{Wenhan Xiong},
  \bibinfo{person}{Hong Wang}, {and} \bibinfo{person}{William Wang}.}
  \bibinfo{year}{2020}\natexlab{}.
\newblock \showarticletitle{Hybridqa: A dataset of multi-hop question answering
  over tabular and textual data}.
\newblock \bibinfo{journal}{\emph{arXiv preprint arXiv:2004.07347}}
  (\bibinfo{year}{2020}).
\newblock


\bibitem[Chhikara et~al\mbox{.}(2025)]%
        {chhikara2025mem0}
\bibfield{author}{\bibinfo{person}{Prateek Chhikara}, \bibinfo{person}{Dev
  Khant}, \bibinfo{person}{Saket Aryan}, \bibinfo{person}{Taranjeet Singh},
  {and} \bibinfo{person}{Deshraj Yadav}.} \bibinfo{year}{2025}\natexlab{}.
\newblock \showarticletitle{Mem0: Building production-ready ai agents with
  scalable long-term memory}.
\newblock \bibinfo{journal}{\emph{arXiv preprint arXiv:2504.19413}}
  (\bibinfo{year}{2025}).
\newblock


\bibitem[Devlin et~al\mbox{.}(2019)]%
        {devlin2019bert}
\bibfield{author}{\bibinfo{person}{Jacob Devlin}, \bibinfo{person}{Ming-Wei
  Chang}, \bibinfo{person}{Kenton Lee}, {and} \bibinfo{person}{Kristina
  Toutanova}.} \bibinfo{year}{2019}\natexlab{}.
\newblock \showarticletitle{Bert: Pre-training of deep bidirectional
  transformers for language understanding}. In
  \bibinfo{booktitle}{\emph{Proceedings of the 2019 conference of the North
  American chapter of the association for computational linguistics: human
  language technologies, volume 1 (long and short papers)}}.
  \bibinfo{pages}{4171--4186}.
\newblock


\bibitem[Dong et~al\mbox{.}(2022)]%
        {dong2022survey}
\bibfield{author}{\bibinfo{person}{Qingxiu Dong}, \bibinfo{person}{Lei Li},
  \bibinfo{person}{Damai Dai}, \bibinfo{person}{Ce Zheng},
  \bibinfo{person}{Jingyuan Ma}, \bibinfo{person}{Rui Li},
  \bibinfo{person}{Heming Xia}, \bibinfo{person}{Jingjing Xu},
  \bibinfo{person}{Zhiyong Wu}, \bibinfo{person}{Tianyu Liu}, {et~al\mbox{.}}}
  \bibinfo{year}{2022}\natexlab{}.
\newblock \showarticletitle{A survey on in-context learning}.
\newblock \bibinfo{journal}{\emph{arXiv preprint arXiv:2301.00234}}
  (\bibinfo{year}{2022}).
\newblock


\bibitem[Du et~al\mbox{.}(2024)]%
        {du2024perltqa}
\bibfield{author}{\bibinfo{person}{Yiming Du}, \bibinfo{person}{Hongru Wang},
  \bibinfo{person}{Zhengyi Zhao}, \bibinfo{person}{Bin Liang},
  \bibinfo{person}{Baojun Wang}, \bibinfo{person}{Wanjun Zhong},
  \bibinfo{person}{Zezhong Wang}, {and} \bibinfo{person}{Kam-Fai Wong}.}
  \bibinfo{year}{2024}\natexlab{}.
\newblock \showarticletitle{PerLTQA: A Personal Long-Term Memory Dataset for
  Memory Classification, Retrieval, and Synthesis in Question Answering}.
\newblock \bibinfo{journal}{\emph{arXiv preprint arXiv:2402.16288}}
  (\bibinfo{year}{2024}).
\newblock


\bibitem[Fan et~al\mbox{.}(2024)]%
        {fan2024survey}
\bibfield{author}{\bibinfo{person}{Wenqi Fan}, \bibinfo{person}{Yujuan Ding},
  \bibinfo{person}{Liangbo Ning}, \bibinfo{person}{Shijie Wang},
  \bibinfo{person}{Hengyun Li}, \bibinfo{person}{Dawei Yin},
  \bibinfo{person}{Tat-Seng Chua}, {and} \bibinfo{person}{Qing Li}.}
  \bibinfo{year}{2024}\natexlab{}.
\newblock \showarticletitle{A survey on rag meeting llms: Towards
  retrieval-augmented large language models}. In
  \bibinfo{booktitle}{\emph{Proceedings of the 30th ACM SIGKDD Conference on
  Knowledge Discovery and Data Mining}}. \bibinfo{pages}{6491--6501}.
\newblock


\bibitem[Guo et~al\mbox{.}(2024)]%
        {guo2024large}
\bibfield{author}{\bibinfo{person}{Taicheng Guo}, \bibinfo{person}{Xiuying
  Chen}, \bibinfo{person}{Yaqi Wang}, \bibinfo{person}{Ruidi Chang},
  \bibinfo{person}{Shichao Pei}, \bibinfo{person}{Nitesh~V Chawla},
  \bibinfo{person}{Olaf Wiest}, {and} \bibinfo{person}{Xiangliang Zhang}.}
  \bibinfo{year}{2024}\natexlab{}.
\newblock \showarticletitle{Large language model based multi-agents: A survey
  of progress and challenges}.
\newblock \bibinfo{journal}{\emph{arXiv preprint arXiv:2402.01680}}
  (\bibinfo{year}{2024}).
\newblock


\bibitem[Ho et~al\mbox{.}(2020)]%
        {ho2020constructing}
\bibfield{author}{\bibinfo{person}{Xanh Ho}, \bibinfo{person}{Anh-Khoa~Duong
  Nguyen}, \bibinfo{person}{Saku Sugawara}, {and} \bibinfo{person}{Akiko
  Aizawa}.} \bibinfo{year}{2020}\natexlab{}.
\newblock \showarticletitle{Constructing a multi-hop qa dataset for
  comprehensive evaluation of reasoning steps}.
\newblock \bibinfo{journal}{\emph{arXiv preprint arXiv:2011.01060}}
  (\bibinfo{year}{2020}).
\newblock


\bibitem[Hu et~al\mbox{.}(2022)]%
        {hu2022lora}
\bibfield{author}{\bibinfo{person}{Edward~J Hu}, \bibinfo{person}{Yelong Shen},
  \bibinfo{person}{Phillip Wallis}, \bibinfo{person}{Zeyuan Allen-Zhu},
  \bibinfo{person}{Yuanzhi Li}, \bibinfo{person}{Shean Wang},
  \bibinfo{person}{Lu Wang}, \bibinfo{person}{Weizhu Chen}, {et~al\mbox{.}}}
  \bibinfo{year}{2022}\natexlab{}.
\newblock \showarticletitle{Lora: Low-rank adaptation of large language
  models.}
\newblock \bibinfo{journal}{\emph{ICLR}} \bibinfo{volume}{1},
  \bibinfo{number}{2} (\bibinfo{year}{2022}), \bibinfo{pages}{3}.
\newblock


\bibitem[Huang et~al\mbox{.}(2024)]%
        {huang2024understanding}
\bibfield{author}{\bibinfo{person}{Xu Huang}, \bibinfo{person}{Weiwen Liu},
  \bibinfo{person}{Xiaolong Chen}, \bibinfo{person}{Xingmei Wang},
  \bibinfo{person}{Hao Wang}, \bibinfo{person}{Defu Lian},
  \bibinfo{person}{Yasheng Wang}, \bibinfo{person}{Ruiming Tang}, {and}
  \bibinfo{person}{Enhong Chen}.} \bibinfo{year}{2024}\natexlab{}.
\newblock \showarticletitle{Understanding the planning of LLM agents: A
  survey}.
\newblock \bibinfo{journal}{\emph{arXiv preprint arXiv:2402.02716}}
  (\bibinfo{year}{2024}).
\newblock


\bibitem[Kumar et~al\mbox{.}(2024)]%
        {kumar2024longlamp}
\bibfield{author}{\bibinfo{person}{Ishita Kumar}, \bibinfo{person}{Snigdha
  Viswanathan}, \bibinfo{person}{Sushrita Yerra}, \bibinfo{person}{Alireza
  Salemi}, \bibinfo{person}{Ryan~A Rossi}, \bibinfo{person}{Franck
  Dernoncourt}, \bibinfo{person}{Hanieh Deilamsalehy}, \bibinfo{person}{Xiang
  Chen}, \bibinfo{person}{Ruiyi Zhang}, \bibinfo{person}{Shubham Agarwal},
  {et~al\mbox{.}}} \bibinfo{year}{2024}\natexlab{}.
\newblock \showarticletitle{Longlamp: A benchmark for personalized long-form
  text generation}.
\newblock \bibinfo{journal}{\emph{arXiv preprint arXiv:2407.11016}}
  (\bibinfo{year}{2024}).
\newblock


\bibitem[Li et~al\mbox{.}(2024)]%
        {li2024personal}
\bibfield{author}{\bibinfo{person}{Yuanchun Li}, \bibinfo{person}{Hao Wen},
  \bibinfo{person}{Weijun Wang}, \bibinfo{person}{Xiangyu Li},
  \bibinfo{person}{Yizhen Yuan}, \bibinfo{person}{Guohong Liu},
  \bibinfo{person}{Jiacheng Liu}, \bibinfo{person}{Wenxing Xu},
  \bibinfo{person}{Xiang Wang}, \bibinfo{person}{Yi Sun}, {et~al\mbox{.}}}
  \bibinfo{year}{2024}\natexlab{}.
\newblock \showarticletitle{Personal llm agents: Insights and survey about the
  capability, efficiency and security}.
\newblock \bibinfo{journal}{\emph{arXiv preprint arXiv:2401.05459}}
  (\bibinfo{year}{2024}).
\newblock


\bibitem[Lloyd(1982)]%
        {lloyd1982least}
\bibfield{author}{\bibinfo{person}{Stuart Lloyd}.}
  \bibinfo{year}{1982}\natexlab{}.
\newblock \showarticletitle{Least squares quantization in PCM}.
\newblock \bibinfo{journal}{\emph{IEEE transactions on information theory}}
  \bibinfo{volume}{28}, \bibinfo{number}{2} (\bibinfo{year}{1982}),
  \bibinfo{pages}{129--137}.
\newblock


\bibitem[Lù(2024)]%
        {bm25s}
\bibfield{author}{\bibinfo{person}{Xing~Han Lù}.}
  \bibinfo{year}{2024}\natexlab{}.
\newblock \bibinfo{title}{BM25S: Orders of magnitude faster lexical search via
  eager sparse scoring}.
\newblock
\showeprint[arxiv]{2407.03618}~[cs.IR]
\urldef\tempurl%
\url{https://arxiv.org/abs/2407.03618}
\showURL{%
\tempurl}


\bibitem[Mitchell et~al\mbox{.}(2021)]%
        {mitchell2021fast}
\bibfield{author}{\bibinfo{person}{Eric Mitchell}, \bibinfo{person}{Charles
  Lin}, \bibinfo{person}{Antoine Bosselut}, \bibinfo{person}{Chelsea Finn},
  {and} \bibinfo{person}{Christopher~D Manning}.}
  \bibinfo{year}{2021}\natexlab{}.
\newblock \showarticletitle{Fast model editing at scale}.
\newblock \bibinfo{journal}{\emph{arXiv preprint arXiv:2110.11309}}
  (\bibinfo{year}{2021}).
\newblock


\bibitem[Naveed et~al\mbox{.}(2023)]%
        {naveed2023comprehensive}
\bibfield{author}{\bibinfo{person}{Humza Naveed}, \bibinfo{person}{Asad~Ullah
  Khan}, \bibinfo{person}{Shi Qiu}, \bibinfo{person}{Muhammad Saqib},
  \bibinfo{person}{Saeed Anwar}, \bibinfo{person}{Muhammad Usman},
  \bibinfo{person}{Naveed Akhtar}, \bibinfo{person}{Nick Barnes}, {and}
  \bibinfo{person}{Ajmal Mian}.} \bibinfo{year}{2023}\natexlab{}.
\newblock \showarticletitle{A comprehensive overview of large language models}.
\newblock \bibinfo{journal}{\emph{ACM Transactions on Intelligent Systems and
  Technology}} (\bibinfo{year}{2023}).
\newblock


\bibitem[Peng et~al\mbox{.}(2024)]%
        {peng2024graph}
\bibfield{author}{\bibinfo{person}{Boci Peng}, \bibinfo{person}{Yun Zhu},
  \bibinfo{person}{Yongchao Liu}, \bibinfo{person}{Xiaohe Bo},
  \bibinfo{person}{Haizhou Shi}, \bibinfo{person}{Chuntao Hong},
  \bibinfo{person}{Yan Zhang}, {and} \bibinfo{person}{Siliang Tang}.}
  \bibinfo{year}{2024}\natexlab{}.
\newblock \showarticletitle{Graph retrieval-augmented generation: A survey}.
\newblock \bibinfo{journal}{\emph{arXiv preprint arXiv:2408.08921}}
  (\bibinfo{year}{2024}).
\newblock


\bibitem[Rajpurkar et~al\mbox{.}(2016)]%
        {rajpurkar2016squad}
\bibfield{author}{\bibinfo{person}{Pranav Rajpurkar}, \bibinfo{person}{Jian
  Zhang}, \bibinfo{person}{Konstantin Lopyrev}, {and} \bibinfo{person}{Percy
  Liang}.} \bibinfo{year}{2016}\natexlab{}.
\newblock \showarticletitle{Squad: 100,000+ questions for machine comprehension
  of text}.
\newblock \bibinfo{journal}{\emph{arXiv preprint arXiv:1606.05250}}
  (\bibinfo{year}{2016}).
\newblock


\bibitem[Rezazadeh et~al\mbox{.}(2024)]%
        {rezazadeh2024isolated}
\bibfield{author}{\bibinfo{person}{Alireza Rezazadeh}, \bibinfo{person}{Zichao
  Li}, \bibinfo{person}{Wei Wei}, {and} \bibinfo{person}{Yujia Bao}.}
  \bibinfo{year}{2024}\natexlab{}.
\newblock \showarticletitle{From Isolated Conversations to Hierarchical
  Schemas: Dynamic Tree Memory Representation for LLMs}.
\newblock \bibinfo{journal}{\emph{arXiv preprint arXiv:2410.14052}}
  (\bibinfo{year}{2024}).
\newblock


\bibitem[Salemi et~al\mbox{.}(2023)]%
        {salemi2023lamp}
\bibfield{author}{\bibinfo{person}{Alireza Salemi}, \bibinfo{person}{Sheshera
  Mysore}, \bibinfo{person}{Michael Bendersky}, {and} \bibinfo{person}{Hamed
  Zamani}.} \bibinfo{year}{2023}\natexlab{}.
\newblock \showarticletitle{Lamp: When large language models meet
  personalization}.
\newblock \bibinfo{journal}{\emph{arXiv preprint arXiv:2304.11406}}
  (\bibinfo{year}{2023}).
\newblock


\bibitem[Schnitzler et~al\mbox{.}(2024)]%
        {schnitzler2024morehopqa}
\bibfield{author}{\bibinfo{person}{Julian Schnitzler}, \bibinfo{person}{Xanh
  Ho}, \bibinfo{person}{Jiahao Huang}, \bibinfo{person}{Florian Boudin},
  \bibinfo{person}{Saku Sugawara}, {and} \bibinfo{person}{Akiko Aizawa}.}
  \bibinfo{year}{2024}\natexlab{}.
\newblock \showarticletitle{Morehopqa: More than multi-hop reasoning}.
\newblock \bibinfo{journal}{\emph{arXiv preprint arXiv:2406.13397}}
  (\bibinfo{year}{2024}).
\newblock


\bibitem[Shi et~al\mbox{.}(2025)]%
        {shi2025retrieval}
\bibfield{author}{\bibinfo{person}{Teng Shi}, \bibinfo{person}{Jun Xu},
  \bibinfo{person}{Xiao Zhang}, \bibinfo{person}{Xiaoxue Zang},
  \bibinfo{person}{Kai Zheng}, \bibinfo{person}{Yang Song}, {and}
  \bibinfo{person}{Han Li}.} \bibinfo{year}{2025}\natexlab{}.
\newblock \showarticletitle{Retrieval Augmented Generation with Collaborative
  Filtering for Personalized Text Generation}.
\newblock \bibinfo{journal}{\emph{arXiv preprint arXiv:2504.05731}}
  (\bibinfo{year}{2025}).
\newblock


\bibitem[Sparck~Jones(1972)]%
        {sparck1972statistical}
\bibfield{author}{\bibinfo{person}{Karen Sparck~Jones}.}
  \bibinfo{year}{1972}\natexlab{}.
\newblock \showarticletitle{A statistical interpretation of term specificity
  and its application in retrieval}.
\newblock \bibinfo{journal}{\emph{Journal of documentation}}
  \bibinfo{volume}{28}, \bibinfo{number}{1} (\bibinfo{year}{1972}),
  \bibinfo{pages}{11--21}.
\newblock


\bibitem[Tan et~al\mbox{.}(2025)]%
        {tan2025membench}
\bibfield{author}{\bibinfo{person}{Haoran Tan}, \bibinfo{person}{Zeyu Zhang},
  \bibinfo{person}{Chen Ma}, \bibinfo{person}{Xu Chen}, \bibinfo{person}{Quanyu
  Dai}, {and} \bibinfo{person}{Zhenhua Dong}.} \bibinfo{year}{2025}\natexlab{}.
\newblock \bibinfo{title}{MemBench: Towards More Comprehensive Evaluation on
  the Memory of LLM-based Agents}.
\newblock
\showeprint[arxiv]{2506.21605}~[cs.CL]
\urldef\tempurl%
\url{https://arxiv.org/abs/2506.21605}
\showURL{%
\tempurl}


\bibitem[Tan et~al\mbox{.}(2024)]%
        {tan2024democratizing}
\bibfield{author}{\bibinfo{person}{Zhaoxuan Tan}, \bibinfo{person}{Qingkai
  Zeng}, \bibinfo{person}{Yijun Tian}, \bibinfo{person}{Zheyuan Liu},
  \bibinfo{person}{Bing Yin}, {and} \bibinfo{person}{Meng Jiang}.}
  \bibinfo{year}{2024}\natexlab{}.
\newblock \showarticletitle{Democratizing large language models via
  personalized parameter-efficient fine-tuning}.
\newblock \bibinfo{journal}{\emph{arXiv preprint arXiv:2402.04401}}
  (\bibinfo{year}{2024}).
\newblock


\bibitem[Tang and Yang(2024)]%
        {tang2024multihop}
\bibfield{author}{\bibinfo{person}{Yixuan Tang} {and} \bibinfo{person}{Yi
  Yang}.} \bibinfo{year}{2024}\natexlab{}.
\newblock \showarticletitle{Multihop-rag: Benchmarking retrieval-augmented
  generation for multi-hop queries}.
\newblock \bibinfo{journal}{\emph{arXiv preprint arXiv:2401.15391}}
  (\bibinfo{year}{2024}).
\newblock


\bibitem[Trivedi et~al\mbox{.}(2022)]%
        {trivedi2022musique}
\bibfield{author}{\bibinfo{person}{Harsh Trivedi}, \bibinfo{person}{Niranjan
  Balasubramanian}, \bibinfo{person}{Tushar Khot}, {and}
  \bibinfo{person}{Ashish Sabharwal}.} \bibinfo{year}{2022}\natexlab{}.
\newblock \showarticletitle{MuSiQue: Multihop Questions via Single-hop Question
  Composition}.
\newblock \bibinfo{journal}{\emph{Transactions of the Association for
  Computational Linguistics}}  \bibinfo{volume}{10} (\bibinfo{year}{2022}),
  \bibinfo{pages}{539--554}.
\newblock


\bibitem[Wang et~al\mbox{.}(2024)]%
        {wang2024survey}
\bibfield{author}{\bibinfo{person}{Lei Wang}, \bibinfo{person}{Chen Ma},
  \bibinfo{person}{Xueyang Feng}, \bibinfo{person}{Zeyu Zhang},
  \bibinfo{person}{Hao Yang}, \bibinfo{person}{Jingsen Zhang},
  \bibinfo{person}{Zhiyuan Chen}, \bibinfo{person}{Jiakai Tang},
  \bibinfo{person}{Xu Chen}, \bibinfo{person}{Yankai Lin}, {et~al\mbox{.}}}
  \bibinfo{year}{2024}\natexlab{}.
\newblock \showarticletitle{A survey on large language model based autonomous
  agents}.
\newblock \bibinfo{journal}{\emph{Frontiers of Computer Science}}
  \bibinfo{volume}{18}, \bibinfo{number}{6} (\bibinfo{year}{2024}),
  \bibinfo{pages}{186345}.
\newblock


\bibitem[Wang et~al\mbox{.}(2023)]%
        {wang2023plan}
\bibfield{author}{\bibinfo{person}{Lei Wang}, \bibinfo{person}{Wanyu Xu},
  \bibinfo{person}{Yihuai Lan}, \bibinfo{person}{Zhiqiang Hu},
  \bibinfo{person}{Yunshi Lan}, \bibinfo{person}{Roy Ka-Wei Lee}, {and}
  \bibinfo{person}{Ee-Peng Lim}.} \bibinfo{year}{2023}\natexlab{}.
\newblock \showarticletitle{Plan-and-solve prompting: Improving zero-shot
  chain-of-thought reasoning by large language models}.
\newblock \bibinfo{journal}{\emph{arXiv preprint arXiv:2305.04091}}
  (\bibinfo{year}{2023}).
\newblock


\bibitem[Wang et~al\mbox{.}(2022)]%
        {wang2022text}
\bibfield{author}{\bibinfo{person}{Liang Wang}, \bibinfo{person}{Nan Yang},
  \bibinfo{person}{Xiaolong Huang}, \bibinfo{person}{Binxing Jiao},
  \bibinfo{person}{Linjun Yang}, \bibinfo{person}{Daxin Jiang},
  \bibinfo{person}{Rangan Majumder}, {and} \bibinfo{person}{Furu Wei}.}
  \bibinfo{year}{2022}\natexlab{}.
\newblock \showarticletitle{Text Embeddings by Weakly-Supervised Contrastive
  Pre-training}.
\newblock \bibinfo{journal}{\emph{arXiv preprint arXiv:2212.03533}}
  (\bibinfo{year}{2022}).
\newblock


\bibitem[Wei et~al\mbox{.}(2022)]%
        {wei2022chain}
\bibfield{author}{\bibinfo{person}{Jason Wei}, \bibinfo{person}{Xuezhi Wang},
  \bibinfo{person}{Dale Schuurmans}, \bibinfo{person}{Maarten Bosma},
  \bibinfo{person}{Fei Xia}, \bibinfo{person}{Ed Chi}, \bibinfo{person}{Quoc~V
  Le}, \bibinfo{person}{Denny Zhou}, {et~al\mbox{.}}}
  \bibinfo{year}{2022}\natexlab{}.
\newblock \showarticletitle{Chain-of-thought prompting elicits reasoning in
  large language models}.
\newblock \bibinfo{journal}{\emph{Advances in neural information processing
  systems}}  \bibinfo{volume}{35} (\bibinfo{year}{2022}),
  \bibinfo{pages}{24824--24837}.
\newblock


\bibitem[Wu et~al\mbox{.}(2024a)]%
        {wu2024longmemeval}
\bibfield{author}{\bibinfo{person}{Di Wu}, \bibinfo{person}{Hongwei Wang},
  \bibinfo{person}{Wenhao Yu}, \bibinfo{person}{Yuwei Zhang},
  \bibinfo{person}{Kai-Wei Chang}, {and} \bibinfo{person}{Dong Yu}.}
  \bibinfo{year}{2024}\natexlab{a}.
\newblock \showarticletitle{Longmemeval: Benchmarking chat assistants on
  long-term interactive memory}.
\newblock \bibinfo{journal}{\emph{arXiv preprint arXiv:2410.10813}}
  (\bibinfo{year}{2024}).
\newblock


\bibitem[Wu et~al\mbox{.}(2024b)]%
        {wu2024cofca}
\bibfield{author}{\bibinfo{person}{Jian Wu}, \bibinfo{person}{Linyi Yang},
  \bibinfo{person}{Zhen Wang}, \bibinfo{person}{Manabu Okumura}, {and}
  \bibinfo{person}{Yue Zhang}.} \bibinfo{year}{2024}\natexlab{b}.
\newblock \showarticletitle{Cofca: A Step-Wise Counterfactual Multi-hop QA
  benchmark}.
\newblock \bibinfo{journal}{\emph{arXiv preprint arXiv:2402.11924}}
  (\bibinfo{year}{2024}).
\newblock


\bibitem[Xi et~al\mbox{.}(2025)]%
        {xi2025rise}
\bibfield{author}{\bibinfo{person}{Zhiheng Xi}, \bibinfo{person}{Wenxiang
  Chen}, \bibinfo{person}{Xin Guo}, \bibinfo{person}{Wei He},
  \bibinfo{person}{Yiwen Ding}, \bibinfo{person}{Boyang Hong},
  \bibinfo{person}{Ming Zhang}, \bibinfo{person}{Junzhe Wang},
  \bibinfo{person}{Senjie Jin}, \bibinfo{person}{Enyu Zhou}, {et~al\mbox{.}}}
  \bibinfo{year}{2025}\natexlab{}.
\newblock \showarticletitle{The rise and potential of large language model
  based agents: A survey}.
\newblock \bibinfo{journal}{\emph{Science China Information Sciences}}
  \bibinfo{volume}{68}, \bibinfo{number}{2} (\bibinfo{year}{2025}),
  \bibinfo{pages}{121101}.
\newblock


\bibitem[Yang et~al\mbox{.}(2024)]%
        {qwen2.5}
\bibfield{author}{\bibinfo{person}{An Yang}, \bibinfo{person}{Baosong Yang},
  \bibinfo{person}{Beichen Zhang}, \bibinfo{person}{Binyuan Hui},
  \bibinfo{person}{Bo Zheng}, \bibinfo{person}{Bowen Yu},
  \bibinfo{person}{Chengyuan Li}, \bibinfo{person}{Dayiheng Liu},
  \bibinfo{person}{Fei Huang}, \bibinfo{person}{Haoran Wei},
  \bibinfo{person}{Huan Lin}, \bibinfo{person}{Jian Yang},
  \bibinfo{person}{Jianhong Tu}, \bibinfo{person}{Jianwei Zhang},
  \bibinfo{person}{Jianxin Yang}, \bibinfo{person}{Jiaxi Yang},
  \bibinfo{person}{Jingren Zhou}, \bibinfo{person}{Junyang Lin},
  \bibinfo{person}{Kai Dang}, \bibinfo{person}{Keming Lu},
  \bibinfo{person}{Keqin Bao}, \bibinfo{person}{Kexin Yang},
  \bibinfo{person}{Le Yu}, \bibinfo{person}{Mei Li}, \bibinfo{person}{Mingfeng
  Xue}, \bibinfo{person}{Pei Zhang}, \bibinfo{person}{Qin Zhu},
  \bibinfo{person}{Rui Men}, \bibinfo{person}{Runji Lin},
  \bibinfo{person}{Tianhao Li}, \bibinfo{person}{Tingyu Xia},
  \bibinfo{person}{Xingzhang Ren}, \bibinfo{person}{Xuancheng Ren},
  \bibinfo{person}{Yang Fan}, \bibinfo{person}{Yang Su},
  \bibinfo{person}{Yichang Zhang}, \bibinfo{person}{Yu Wan},
  \bibinfo{person}{Yuqiong Liu}, \bibinfo{person}{Zeyu Cui},
  \bibinfo{person}{Zhenru Zhang}, {and} \bibinfo{person}{Zihan Qiu}.}
  \bibinfo{year}{2024}\natexlab{}.
\newblock \showarticletitle{Qwen2.5 Technical Report}.
\newblock \bibinfo{journal}{\emph{arXiv preprint arXiv:2412.15115}}
  (\bibinfo{year}{2024}).
\newblock


\bibitem[Yang et~al\mbox{.}(2018)]%
        {yang2018hotpotqa}
\bibfield{author}{\bibinfo{person}{Zhilin Yang}, \bibinfo{person}{Peng Qi},
  \bibinfo{person}{Saizheng Zhang}, \bibinfo{person}{Yoshua Bengio},
  \bibinfo{person}{William~W Cohen}, \bibinfo{person}{Ruslan Salakhutdinov},
  {and} \bibinfo{person}{Christopher~D Manning}.}
  \bibinfo{year}{2018}\natexlab{}.
\newblock \showarticletitle{HotpotQA: A dataset for diverse, explainable
  multi-hop question answering}.
\newblock \bibinfo{journal}{\emph{arXiv preprint arXiv:1809.09600}}
  (\bibinfo{year}{2018}).
\newblock


\bibitem[Yao et~al\mbox{.}(2023)]%
        {yao2023tree}
\bibfield{author}{\bibinfo{person}{Shunyu Yao}, \bibinfo{person}{Dian Yu},
  \bibinfo{person}{Jeffrey Zhao}, \bibinfo{person}{Izhak Shafran},
  \bibinfo{person}{Tom Griffiths}, \bibinfo{person}{Yuan Cao}, {and}
  \bibinfo{person}{Karthik Narasimhan}.} \bibinfo{year}{2023}\natexlab{}.
\newblock \showarticletitle{Tree of thoughts: Deliberate problem solving with
  large language models}.
\newblock \bibinfo{journal}{\emph{Advances in neural information processing
  systems}}  \bibinfo{volume}{36} (\bibinfo{year}{2023}),
  \bibinfo{pages}{11809--11822}.
\newblock


\bibitem[Zhang et~al\mbox{.}(2025b)]%
        {zhang2025survey_tts}
\bibfield{author}{\bibinfo{person}{Qiyuan Zhang}, \bibinfo{person}{Fuyuan Lyu},
  \bibinfo{person}{Zexu Sun}, \bibinfo{person}{Lei Wang},
  \bibinfo{person}{Weixu Zhang}, \bibinfo{person}{Wenyue Hua},
  \bibinfo{person}{Haolun Wu}, \bibinfo{person}{Zhihan Guo},
  \bibinfo{person}{Yufei Wang}, \bibinfo{person}{Niklas Muennighoff},
  {et~al\mbox{.}}} \bibinfo{year}{2025}\natexlab{b}.
\newblock \showarticletitle{A Survey on Test-Time Scaling in Large Language
  Models: What, How, Where, and How Well?}
\newblock \bibinfo{journal}{\emph{arXiv preprint arXiv:2503.24235}}
  (\bibinfo{year}{2025}).
\newblock


\bibitem[Zhang et~al\mbox{.}(2024c)]%
        {zhang2024instruction}
\bibfield{author}{\bibinfo{person}{Shengyu Zhang}, \bibinfo{person}{Linfeng
  Dong}, \bibinfo{person}{Xiaoya Li}, \bibinfo{person}{Sen Zhang},
  \bibinfo{person}{Xiaofei Sun}, \bibinfo{person}{Shuhe Wang},
  \bibinfo{person}{Jiwei Li}, \bibinfo{person}{Runyi Hu},
  \bibinfo{person}{Tianwei Zhang}, \bibinfo{person}{Fei Wu}, {and}
  \bibinfo{person}{Guoyin Wang}.} \bibinfo{year}{2024}\natexlab{c}.
\newblock \bibinfo{title}{Instruction Tuning for Large Language Models: A
  Survey}.
\newblock
\showeprint[arxiv]{2308.10792}~[cs.CL]
\urldef\tempurl%
\url{https://arxiv.org/abs/2308.10792}
\showURL{%
\tempurl}


\bibitem[Zhang et~al\mbox{.}(2024a)]%
        {zhang2024survey}
\bibfield{author}{\bibinfo{person}{Zeyu Zhang}, \bibinfo{person}{Xiaohe Bo},
  \bibinfo{person}{Chen Ma}, \bibinfo{person}{Rui Li}, \bibinfo{person}{Xu
  Chen}, \bibinfo{person}{Quanyu Dai}, \bibinfo{person}{Jieming Zhu},
  \bibinfo{person}{Zhenhua Dong}, {and} \bibinfo{person}{Ji-Rong Wen}.}
  \bibinfo{year}{2024}\natexlab{a}.
\newblock \showarticletitle{A survey on the memory mechanism of large language
  model based agents}.
\newblock \bibinfo{journal}{\emph{arXiv preprint arXiv:2404.13501}}
  (\bibinfo{year}{2024}).
\newblock


\bibitem[Zhang et~al\mbox{.}(2024b)]%
        {zhang2024memsim}
\bibfield{author}{\bibinfo{person}{Zeyu Zhang}, \bibinfo{person}{Quanyu Dai},
  \bibinfo{person}{Luyu Chen}, \bibinfo{person}{Zeren Jiang},
  \bibinfo{person}{Rui Li}, \bibinfo{person}{Jieming Zhu}, \bibinfo{person}{Xu
  Chen}, \bibinfo{person}{Yi Xie}, \bibinfo{person}{Zhenhua Dong}, {and}
  \bibinfo{person}{Ji-Rong Wen}.} \bibinfo{year}{2024}\natexlab{b}.
\newblock \showarticletitle{Memsim: A bayesian simulator for evaluating memory
  of llm-based personal assistants}.
\newblock \bibinfo{journal}{\emph{arXiv preprint arXiv:2409.20163}}
  (\bibinfo{year}{2024}).
\newblock


\bibitem[Zhang et~al\mbox{.}(2025a)]%
        {zhang2025memengine}
\bibfield{author}{\bibinfo{person}{Zeyu Zhang}, \bibinfo{person}{Quanyu Dai},
  \bibinfo{person}{Xu Chen}, \bibinfo{person}{Rui Li},
  \bibinfo{person}{Zhongyang Li}, {and} \bibinfo{person}{Zhenhua Dong}.}
  \bibinfo{year}{2025}\natexlab{a}.
\newblock \showarticletitle{MemEngine: A Unified and Modular Library for
  Developing Advanced Memory of LLM-based Agents}. In
  \bibinfo{booktitle}{\emph{Companion Proceedings of the ACM on Web Conference
  2025}}. \bibinfo{pages}{821--824}.
\newblock


\bibitem[Zhao et~al\mbox{.}(2023)]%
        {zhao2023survey}
\bibfield{author}{\bibinfo{person}{Wayne~Xin Zhao}, \bibinfo{person}{Kun Zhou},
  \bibinfo{person}{Junyi Li}, \bibinfo{person}{Tianyi Tang},
  \bibinfo{person}{Xiaolei Wang}, \bibinfo{person}{Yupeng Hou},
  \bibinfo{person}{Yingqian Min}, \bibinfo{person}{Beichen Zhang},
  \bibinfo{person}{Junjie Zhang}, \bibinfo{person}{Zican Dong},
  {et~al\mbox{.}}} \bibinfo{year}{2023}\natexlab{}.
\newblock \showarticletitle{A survey of large language models}.
\newblock \bibinfo{journal}{\emph{arXiv preprint arXiv:2303.18223}}
  \bibinfo{volume}{1}, \bibinfo{number}{2} (\bibinfo{year}{2023}).
\newblock


\bibitem[Zhong et~al\mbox{.}(2024)]%
        {zhong2024memorybank}
\bibfield{author}{\bibinfo{person}{Wanjun Zhong}, \bibinfo{person}{Lianghong
  Guo}, \bibinfo{person}{Qiqi Gao}, \bibinfo{person}{He Ye}, {and}
  \bibinfo{person}{Yanlin Wang}.} \bibinfo{year}{2024}\natexlab{}.
\newblock \showarticletitle{Memorybank: Enhancing large language models with
  long-term memory}. In \bibinfo{booktitle}{\emph{Proceedings of the AAAI
  Conference on Artificial Intelligence}}, Vol.~\bibinfo{volume}{38}.
  \bibinfo{pages}{19724--19731}.
\newblock


\bibitem[Zhong et~al\mbox{.}(2023)]%
        {zhong2023mquake}
\bibfield{author}{\bibinfo{person}{Zexuan Zhong}, \bibinfo{person}{Zhengxuan
  Wu}, \bibinfo{person}{Christopher~D Manning}, \bibinfo{person}{Christopher
  Potts}, {and} \bibinfo{person}{Danqi Chen}.} \bibinfo{year}{2023}\natexlab{}.
\newblock \showarticletitle{Mquake: Assessing knowledge editing in language
  models via multi-hop questions}.
\newblock \bibinfo{journal}{\emph{arXiv preprint arXiv:2305.14795}}
  (\bibinfo{year}{2023}).
\newblock


\bibitem[Zhu et~al\mbox{.}(2024)]%
        {zhu2024fanoutqa}
\bibfield{author}{\bibinfo{person}{Andrew Zhu}, \bibinfo{person}{Alyssa Hwang},
  \bibinfo{person}{Liam Dugan}, {and} \bibinfo{person}{Chris Callison-Burch}.}
  \bibinfo{year}{2024}\natexlab{}.
\newblock \showarticletitle{Fanoutqa: Multi-hop, multi-document question
  answering for large language models}.
\newblock \bibinfo{journal}{\emph{arXiv preprint arXiv:2402.14116}}
  (\bibinfo{year}{2024}).
\newblock


\end{thebibliography}

\clearpage

\appendix

\section{More Results of Experiments}

\subsection{Results of Implicit Memory with Different Training Epochs}
\label{appendix:sft_steps}
The results are presented as follows:
\begin{figure}[h]
	\centering
	\begin{subfigure}{\linewidth}
		\includegraphics[width=1.0\textwidth]{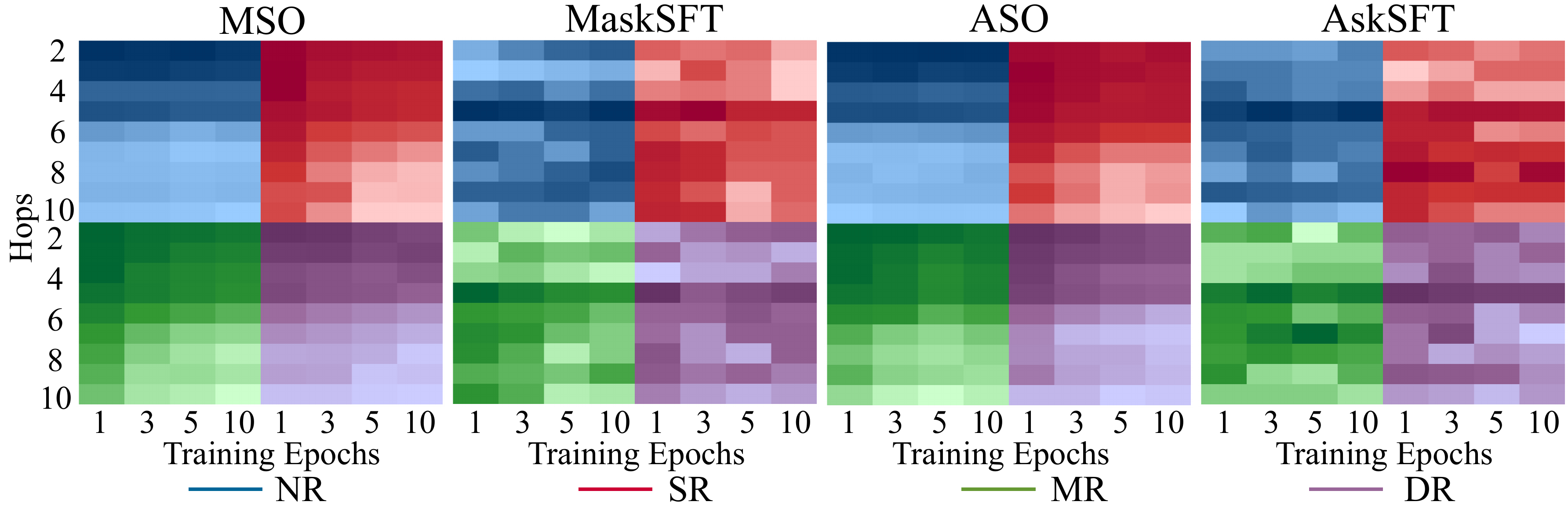}
	\end{subfigure}
	
	\vspace{-0.3cm}
	\caption{Performance of various training epochs.}
	\label{fig:sft_steps}
	\vspace{-0.6cm}
\end{figure}

The results indicate that implicit memory alone cannot achieve great performance even after training for more epochs.
Additionally, the performance of MSO and ASO significantly declines as training steps increase, possibly because more training steps lead to greater reasoning capability degradation.

\subsection{Results of Implicit Memory with Different Reasoning Steps}
\label{appendix:sft_resteps}
The results are presented as follows:
\begin{figure}[h]
	\centering
	\begin{subfigure}{\linewidth}
		\includegraphics[width=1.0\textwidth]{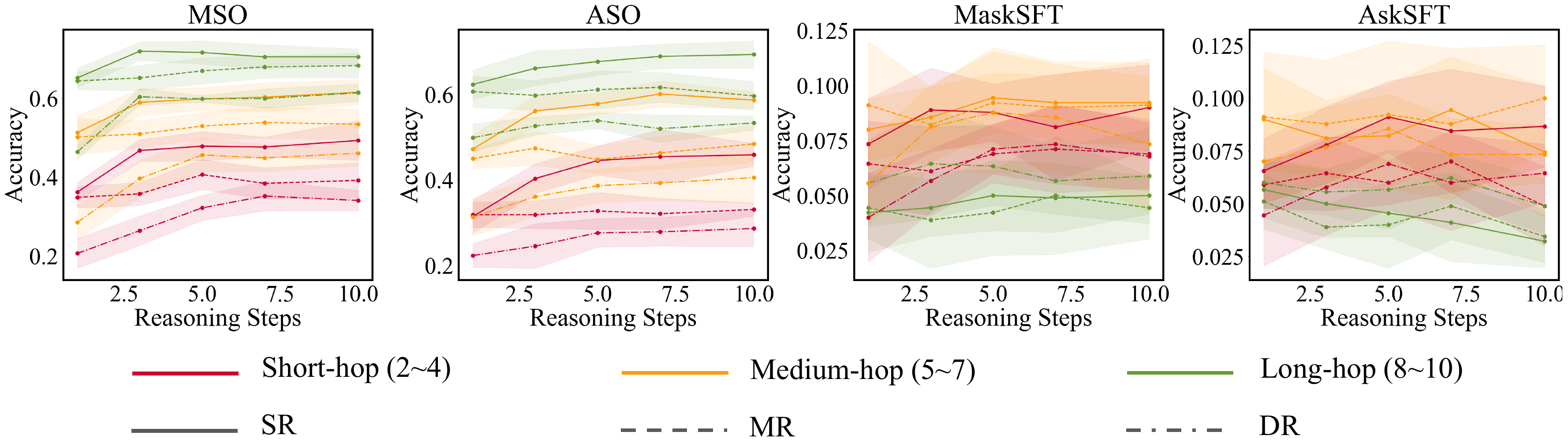}
	\end{subfigure}
	
	\vspace{-0.3cm}
	\caption{Performance of various reasoning steps.}
	\label{fig:sft_resteps}
	\vspace{-0.6cm}
\end{figure}

We find that even with increased reasoning steps, implicit memory cannot achieve reasonable results under our setting. For MSO and ASO, their results exhibit similar patterns to explicit memory as analyzed in the previous section.

\subsection{Results of Implicit Memory with Different Backbone Size}
\label{appendix:sft_backbone}
The results are presented as follows:
\begin{figure}[h]
	\centering
	\begin{subfigure}{\linewidth}
		\includegraphics[width=1.0\textwidth]{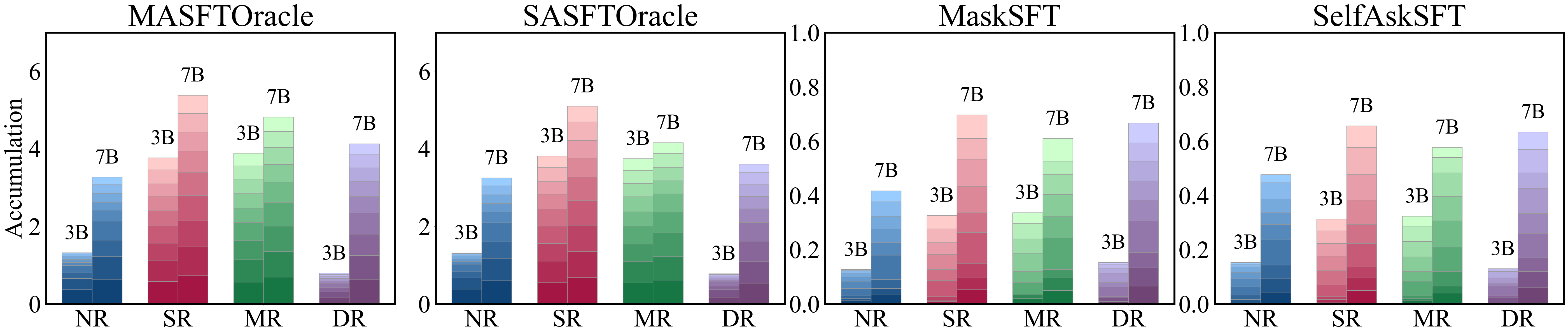}
	\end{subfigure}
	
	\vspace{-0.3cm}
	\caption{Performance of various backbone size.}
	\label{fig:sft_backbone}
	\vspace{-0.6cm}
\end{figure}

Across various reasoning structures and memory baselines, the 3B model shows significant degradation in accuracy.
Among different reasoning structures, NR and DR exhibit the most pronounced performance decline, significantly higher than the other two reasoning structures, with DR declining by approximately 80\%.

\section{More Details of Reasoning Structures}
\label{appendix:reasoning_structures}

\subsection{Prompts of Naive Reasoning}
	The prompts of naive reasoning are as follows:

\begin{tcolorbox}[colback=nr-blue!5!white,colframe=nr-blue!80!white,arc=0.1cm, title={The prompt of NR with explicit memory.}]
	Please help me answer the following question based on the given information.\\
	Information: \textbf{[Retrieved References]}\\
	Question: \textbf{[Question]}\\
	Requirements:\\
	1. your answer should be as concise as possible, commonly in a few words.\\
	2. if the answer is a date, please output it in YYYY-MM-DD format.\\
	3. if the answer is a number, please do not include commas. If the numerical answer has units, please indicate them in parentheses, such as 5 (USD).\\
	You should only output the answer in one line (no code block), without any other descriptions.
\end{tcolorbox}

\begin{tcolorbox}[colback=nr-blue!5!white,colframe=nr-blue!80!white,arc=0.1cm, title={The prompt of NR with implicit memory.}]
	Please help me answer the following question.\\
	Question: \textbf{[Question]}\\
	Requirements:\\
	1. your answer should be as concise as possible, commonly in a few words.\\
	2. if the answer is a date, please output it in YYYY-MM-DD format.\\
	3. if the answer is a number, please do not include commas. If the numerical answer has units, please indicate them in parentheses, such as 5 (USD).\\
	You should only output the answer in one line (no code block), without any other descriptions.
\end{tcolorbox}

\subsection{Prompts of Sequential Reasoning}
The process of sequential reasoning has three parts, including starting, thinking, and answering. The prompts of these steps are presented as follows:
\begin{tcolorbox}[colback=sr-red!5!white,colframe=sr-red!80!white,arc=0.1cm, title={(Starting) The prompt of SR with explicit memory.}]
	To better answer the following question, let's think step by step.\\
	The current step is 1, and you should provide the final answer at step \textbf{[Max Steps]}.\\
	Please generate your thoughts for the current step, and you may refer to the given information.\\
	Information: \textbf{[Retrieved References]} \\
	Question: \textbf{[Question]}\\
	Requirements:\\
	1. Your thoughts should be concise but informative sentences.\\
	2. You should only output the thoughts in one line (no code block).
\end{tcolorbox}

\begin{tcolorbox}[colback=sr-red!5!white,colframe=sr-red!80!white,arc=0.1cm, title={(Thinking) The prompt of SR with explicit memory.}]
	To better answer the following questions, let's think step by step.\\
	The current step is \textbf{[Current Step]}, and you should provide the final answer at step \textbf{[Max Steps]}.\\
	Please generate your thoughts for the current step, and you may refer to the given information and your previous thought.\\
	Information: \textbf{[Retrieved References]} \\
	Previous Thoughts: \textbf{[Previous Thought]}\\
	Question: \textbf{[Question]}\\
	Requirements:\\
	1. Your thoughts should be concise but informative sentences.\\
	2. You should only output the thoughts in one line (no code block).
\end{tcolorbox}

\begin{tcolorbox}[colback=sr-red!5!white,colframe=sr-red!80!white,arc=0.1cm, title={(Answering) The prompt of SR with explicit memory.}]
	To better answer the following question, let's think step by step.\\
	Please help me generate the answer to the question based on the given information and your previous thoughts.\\
	Information: \textbf{[Retrieved References]} \\
	Previous Thoughts: \textbf{[Previous Thought]}\\
	Question: \textbf{[Question]}\\
	Requirements:\\
	1. your answer should be as concise as possible, commonly in a few words.\\
	2. if the answer is a date, please output it in YYYY-MM-DD format.\\
	3. if the answer is a number, please do not include commas. If the numerical answer has units, please indicate them in parentheses, such as 5 (USD).\\
	You should only output the answer in one line (no code block), without any other descriptions.
\end{tcolorbox}

\begin{tcolorbox}[colback=sr-red!5!white,colframe=sr-red!80!white,arc=0.1cm, title={(Starting) The prompt of SR with implicit memory.}]
	To better answer the following question, let's think step by step.\\
	The current step is 1, and you should provide the final answer at step \textbf{[Max Steps]}.\\
	Please generate your thoughts for the current step.\\
	Question: \textbf{[Question]}\\
	Requirements:\\
	1. Your thoughts should be concise but informative sentences.\\
	2. You should only output the thoughts in one line (no code block).
\end{tcolorbox}

\begin{tcolorbox}[colback=sr-red!5!white,colframe=sr-red!80!white,arc=0.1cm, title={(Thinking) The prompt of SR with implicit memory.}]
	To better answer the following questions, let's think step by step.\\
	The current step is \textbf{[Current Step]}, and you should provide the final answer at step \textbf{[Max Steps]}.\\
	Please generate your thoughts for the current step, and you may refer to your previous thought.\\
	Previous Thoughts: \textbf{[Previous Thought]}\\
	Question: \textbf{[Question]}\\
	Requirements:\\
	1. Your thoughts should be concise but informative sentences.\\
	2. You should only output the thoughts in one line (no code block).
\end{tcolorbox}

\begin{tcolorbox}[colback=sr-red!5!white,colframe=sr-red!80!white,arc=0.1cm, title={(Answering) The prompt of SR with implicit memory.}]
	To better answer the following question, let's think step by step.\\
	Please help me generate the answer to the question based on your previous thoughts.\\
	Previous Thoughts: \textbf{[Previous Thought]}\\
	Question: \textbf{[Question]}\\
	Requirements:\\
	1. your answer should be as concise as possible, commonly in a few words.\\
	2. if the answer is a date, please output it in YYYY-MM-DD format.\\
	3. if the answer is a number, please do not include commas. If the numerical answer has units, please indicate them in parentheses, such as 5 (USD).\\
	You should only output the answer in one line (no code block), without any other descriptions.
\end{tcolorbox}

\subsection{Prompts of Multi-path Reasoning}
The process of multi-path reasoning has three parts, including starting, thinking, and answering. The prompts of these steps are presented as follows:
\begin{tcolorbox}[colback=mr-green!5!white,colframe=mr-green!80!white,arc=0.1cm, title={(Starting) The prompt of MR with explicit memory.}]
	To better answer the following question, let's think step by step.\\
	Please generate your thoughts for the current step, and you may refer to the given information.\\
	Information: \textbf{[Retrieved References]} \\
	Question: \textbf{[Question]}\\
	Requirements:\\
	1. Your thoughts should be concise but informative sentences.\\
	2. You should only output the thoughts in one line (no code block).
\end{tcolorbox}

\begin{tcolorbox}[colback=mr-green!5!white,colframe=mr-green!80!white,arc=0.1cm, title={(Thinking) The prompt of MR with explicit memory.}]
	To better answer the following questions, let's think step by step.\\
	Please generate your thoughts for the current step, and you may refer to the given information and your previous thoughts.\\
	Information: \textbf{[Retrieved References]} \\
	Previous Thoughts: \textbf{[Previous Thought]}\\
	Question: \textbf{[Question]}\\
	Requirements:\\
	1. Your thoughts should be concise but informative sentences.\\
	2. You should only output the thoughts in one line (no code block).
\end{tcolorbox}

\begin{tcolorbox}[colback=mr-green!5!white,colframe=mr-green!80!white,arc=0.1cm, title={(Answering) The prompt of MR with explicit memory.}]
	To better answer the following question, let's think step by step.\\
	Please help me generate the answer to the question based on the given information and your previous thoughts.\\
	Information: \textbf{[Retrieved References]} \\
	Previous Thoughts: \textbf{[Previous Thought]}\\
	Question: \textbf{[Question]}\\
	Requirements:\\
	1. your answer should be as concise as possible, commonly in a few words.\\
	2. if the answer is a date, please output it in YYYY-MM-DD format.\\
	3. if the answer is a number, please do not include commas. If the numerical answer has units, please indicate them in parentheses, such as 5 (USD).\\
	You should only output the answer in one line (no code block), without any other descriptions.
\end{tcolorbox}

\begin{tcolorbox}[colback=mr-green!5!white,colframe=mr-green!80!white,arc=0.1cm, title={(Starting) The prompt of MR with implicit memory.}]
	To better answer the following question, let's think step by step.\\
	Please generate your thoughts for the current step.\\
	Question: \textbf{[Question]}\\
	Requirements:\\
	1. Your thoughts should be concise but informative sentences.\\
	2. You should only output the thoughts in one line (no code block).
\end{tcolorbox}

\begin{tcolorbox}[colback=mr-green!5!white,colframe=mr-green!80!white,arc=0.1cm, title={(Thinking) The prompt of MR with implicit memory.}]
	To better answer the following questions, let's think step by step.\\
	Please generate your thoughts for the current step, and you may refer to your previous thoughts.\\
	Previous Thoughts: \textbf{[Previous Thought]}\\
	Question: \textbf{[Question]}\\
	Requirements:\\
	1. Your thoughts should be concise but informative sentences.\\
	2. You should only output the thoughts in one line (no code block).
\end{tcolorbox}

\begin{tcolorbox}[colback=mr-green!5!white,colframe=mr-green!80!white,arc=0.1cm, title={(Answering) The prompt of MR with implicit memory.}]
	To better answer the following question, let's think step by step.\\
	Please help me generate the answer to the question based on your previous thoughts.\\
	Information: \textbf{[Retrieved References]} \\
	Previous Thoughts: \textbf{[Previous Thought]}\\
	Question: \textbf{[Question]}\\
	Requirements:\\
	1. your answer should be as concise as possible, commonly in a few words.\\
	2. if the answer is a date, please output it in YYYY-MM-DD format.\\
	3. if the answer is a number, please do not include commas. If the numerical answer has units, please indicate them in parentheses, such as 5 (USD).\\
	You should only output the answer in one line (no code block), without any other descriptions.
\end{tcolorbox}

\subsection{Prompts of Decomposition Reasoning}
The process of decomposition reasoning has three parts, including dividing, solving, and merging. The prompts of these steps are presented as follows:

\begin{tcolorbox}[colback=dr-purple!5!white,colframe=dr-purple!80!white,arc=0.1cm, title={(Dividing) The prompt of NR with explicit memory.}]
	To better answer the following question, please break it down into several sub-questions.\\
	Question: \textbf{[Question]}\\
	Requirements:\\
	1. The sub-questions should be concise.\\
	2. Each sub-questions is on a separate line, without any other descriptions.\\
	3. You can decompose the problem into 1 to \textbf{[Max Sub-question]} sub-questions, and any content beyond 5 lines will be ignored.
\end{tcolorbox}

\begin{tcolorbox}[colback=dr-purple!5!white,colframe=dr-purple!80!white,arc=0.1cm, title={(Solving) The prompt of NR with explicit memory.}]
	In order to better answer the following question, we have decomposed them into several sub-questions.\\
	Please help me generate the answer to the current sub-question based on the given information.\\
	Question: \textbf{[Question]}\\
	Sub-questions: \textbf{[Current State]}\\
	Current Sub-question: \textbf{[Current Sub-question]} \\
	Information: \textbf{[Retrieved References]} \\
	Requirements:\\
	1. The answer should be concise but informative.\\
	2. You should only output the answer in one line (no code block), without any other descriptions.
\end{tcolorbox}

\begin{tcolorbox}[colback=dr-purple!5!white,colframe=dr-purple!80!white,arc=0.1cm, title={(Merging) The prompt of NR with explicit memory.}]
	In order to better answer the following question, we have decomposed them into several sub-questions and answered them separately.\\
	Please help me generate the final answer to the question based on the sub-questions and the given information.\\
	Question: \textbf{[Question]}\\
	Sub-questions and Answers: \textbf{[Current State]}\\
	Information: \textbf{[Retrieved References]} \\
	Requirements:\\
	1. your answer should be as concise as possible, commonly in a few words.\\
	2. if the answer is a date, please output it in YYYY-MM-DD format.\\
	3. if the answer is a number, please do not include commas. If the numerical answer has units, please indicate them in parentheses, such as 5 (USD).\\
	You should only output the answer in one line (no code block), without any other descriptions.
\end{tcolorbox}

\begin{tcolorbox}[colback=dr-purple!5!white,colframe=dr-purple!80!white,arc=0.1cm, title={(Dividing) The prompt of NR with implicit memory.}]
	To better answer the following question, please break it down into several sub-questions.\\
	Question: \textbf{[Question]}\\
	Requirements:\\
	1. The sub-questions should be concise.\\
	2. Each sub-questions is on a separate line, without any other descriptions.\\
	3. You can decompose the problem into 1 to \textbf{[Max Sub-question]} sub-questions, and any content beyond 5 lines will be ignored.
\end{tcolorbox}

\begin{tcolorbox}[colback=dr-purple!5!white,colframe=dr-purple!80!white,arc=0.1cm, title={(Solving) The prompt of NR with implicit memory.}]
	In order to better answer the following question, we have decomposed them into several sub-questions.\\
	Please help me generate the answer to the current sub-question.\\
	Question: \textbf{[Question]}\\
	Sub-questions: \textbf{[Current State]}\\
	Current Sub-question: \textbf{[Current Sub-question]} \\
	Requirements:\\
	1. The answer should be concise but informative.\\
	2. You should only output the answer in one line (no code block), without any other descriptions.
\end{tcolorbox}

\begin{tcolorbox}[colback=dr-purple!5!white,colframe=dr-purple!80!white,arc=0.1cm, title={(Merging) The prompt of NR with implicit memory.}]
	In order to better answer the following question, we have decomposed them into several sub-questions and answered them separately.\\
	Please help me generate the final answer to the question based on the sub-questions.\\
	Question: \textbf{[Question]}\\
	Sub-questions and Answers: \textbf{[Current State]}\\
	Requirements:\\
	1. your answer should be as concise as possible, commonly in a few words.\\
	2. if the answer is a date, please output it in YYYY-MM-DD format.\\
	3. if the answer is a number, please do not include commas. If the numerical answer has units, please indicate them in parentheses, such as 5 (USD).\\
	You should only output the answer in one line (no code block), without any other descriptions.
\end{tcolorbox}

\end{document}